\documentclass[final,5p,times,twocolumn]{elsarticle}

\usepackage{macros-arxiv}
\newcolumntype{Y}{>{\centering\arraybackslash}X}

\begin{document}
\begin{frontmatter}
%%%%%%%%%%%%%%%%%%%%%%%%%%%%%%%%%%%%%%%%%%%%%%%%%%%%%%%%%%%%%%%%%
\title{The KnowWhereGraph Ontology}
%%%%%%%%%%%%%%%%%%%%%%%%%%%%%%%%%%%%%%%%%%%%%%%%%%%%%%%%%%%%%%%%%
\author[wsu]{Cogan Shimizu}
\ead{cogan.shimizu@wright.edu}

\author[ucsb]{Shirly Stephen}
\ead{shirlystephen@ucsb.edu}

\author[ksu]{Adrita Barua}
\ead{adrita@ksu.edu}

\author[ucsb]{Ling Cai}
\ead{lingcai@ucsb.edu}

\author[wsu]{Antrea Christou}
\ead{christou.2@wright.edu}

\author[ucsb]{Kitty Currier}
\ead{kcurrier@ucsb.edu}

\author[ksu]{Abhilekha Dalal}
\ead{adalal@ksu.edu}

\author[hyd]{Colby K. Fisher}
\ead{colby@hydronoslabs.com}

\author[ksu]{Pascal Hitzler}
\ead{hitzler@ksu.edu}

\author[vie,ucsb]{Krzysztof Janowicz}
\ead{krzysztof.janowicz@univie.ac.at}

\author[asu]{Wenwen Li}
\ead{wenwen@asu.edu}

\author[vie,ucsb]{Zilong Liu}
\ead{zilong.liu@univie.ac.at}

\author[ksu]{Mohammad Saeid Mahdavinejad}
\ead{saeid@ksu.edu}

\author[ut]{Gengchen Mai}
\ead{gengchen.mai@austin.utexas.edu}

\author [msu]{Dean Rehberger}
\ead{rehberge@msu.edu}

\author[nceas]{Mark Schildhauer}
\ead{schild@nceas.ucsb.edu}

\author[vie,ucsb]{Meilin Shi}
\ead{meilin.shi@univie.ac.at}

\author[ksu]{Sanaz Saki Norouzi}
\ead{sanazsn@ksu.edu}

\author[asu]{Yuanyuan Tian}
\ead{yuanyuantian@asu.edu}

\author[asu]{Sizhe Wang}
\ead{wsizhe@asu.edu}

\author[ucsb]{Zhangyu Wang}
\ead{zhangyuwang@ucsb.edu}

\author[ksu]{Joseph Zalewski}
\ead{jzalewski@ksu.edu}

\author[ksu]{Lu Zhou}
\ead{lu.zhou@ksu.edu}

\author[bristol]{Rui Zhu}
\ead{rui.zhu@bristol.ac.uk}

\address[wsu]{Wright State University}
\address[ucsb]{University of California, Santa Barbara}
\address[ksu]{Kansas State University}
\address[hyd]{Hydronos Labs}
\address[vie]{University of Vienna}
\address[asu]{Arizona State University}
\address[ut]{University of Texas at Austin}
\address[msu]{Michigan State University}
\address[nceas]{National Center for Ecological Anaylsis \& Synthesis}
\address[bristol]{University of Bristol}

\cortext[corauthor]{Corresponding author}
\cortext[contrib]{Marked authors contributed equally, all others alphabetical}

%%%%%%%%%%%%%%%%%%%%%%%%%%%%%%%%%%%%%%%%%%%%%%%%%%%%%%%%%%%%%%%%%
% Abstract
\begin{abstract}
    KnowWhereGraph is one of the largest fully publicly available geospatial knowledge graphs. It includes data from 30 layers on natural hazards (e.g., hurricanes, wildfires), climate variables (e.g., air temperature, precipitation), soil properties, crop and land-cover types, demographics, and human health, various place and region identifiers, among other themes. These have been leveraged through the graph by a variety of applications to address challenges in food security and agricultural supply chains; sustainability related to soil conservation practices and farm labor; and delivery of emergency humanitarian aid following a disaster. In this paper, we introduce the ontology that acts as the schema for KnowWhereGraph. This broad overview provides insight into the requirements and design specifications for the graph and its schema, including the development methodology (modular ontology modeling) and the resources utilized to implement, materialize, and deploy KnowWhereGraph with its end-user interfaces and public query SPARQL endpoint.
\end{abstract}

\begin{keyword}
    modular ontology modeling \sep ontology \sep spatially enabled knowledge graphs
\end{keyword}

\end{frontmatter}
%%%%%%%%%%%%%%%%%%%%%%%%%%%%%%%%%%%%%%%%%%%%%%%%%%%%%%%%%%%%%%%%%
%%%%%%%%%%%%%%%%%%%%%%%%%%%%%%%%%%%%%%%%%%%%%%%%%%%%%%%%%%%%%%%%%
\section{Introduction}
\label{sec:intro}
%%%%%
%%% Opening
% Catchy KWG
KnowWhereGraph\footnote{https://knowwheregraph.org/} (KWG) is one of the largest publicly available geospatial knowledge graphs in the world \cite{kwg-aimag-22}.
% Statistics
It brings together over 30 data layers related to observations of natural hazards (e.g., hurricanes, wildfires, and smoke plumes), spatial characteristics related to climate (e.g., temperature, precipitation, and air quality), soil properties, crop and land-cover types, demographics, and human health (e.g., social vulnerability index), among others, resulting in a knowledge graph with over \replace{16} 28 billion triples. 

% Use-case
KWG supports applications in the food, agriculture, and humanitarian relief sectors, and their corresponding supply chains, generally; environmental policy issues relative to interactions among agricultural sustainability, soil conservation practice, and farm labor; and delivery of emergency humanitarian aid within the US and internationally. 

KnowWhereGraph's mission is to provide area briefings within seconds for any region on Earth (currently mostly in the US) to answer questions such as

\begin{itemize}
    \item "What is here?"
    \item "What happened here before?"
    \item "Who [experts] knows more?"
    \item "How does it compare to other regions or previous events?"
\end{itemize}

In order to assist decision-makers and data scientists in enriching their own data with billions of connected, up-to-date contextual information about humans and their environment to rapidly gain the situational awareness required for proper decision-making.

The highly varied but critical importance of the use cases combined with a large number of data layers and triples created unique challenges for the project in terms of management and execution, challenges that required new approaches to structuring the knowledge graph for efficient querying and use.

% Motivation
To this end and to materialize KnowWhereGraph, we have added an additional layer --- a schema --- which is described formally as an ontology. 
% Transition
This allows us to address the design and functionality requirements resulting from the nature of our mission and use cases \cite{diverse-data}.

%%% Design and Requirements Specification
\textbf{Enable Spatial Integration}\quad The primary purpose of KnowWhereGraph is to provide a convenient method for integrating data along a spatiotemporal dimension. This is integral to the mission of the project and, subsequently, a core requirement for the graph and its schema.

\textbf{Facilitate Data Integration} KnowWhereGraph must be capable of providing an overarching framework for the semantic alignment of key terms and concepts \add{through observable properties of features of interest}.

\textbf{Provide Rich Inferences}\quad Beyond a flat representation of the (integrated) datasets, KnowWhereGraph's schema must be expressive enough to infer latent relationships between data layers, such as causality of events or the inheritance of spatial characteristics. Furthermore, it must provide a framework for additional inferences in the future.

\textbf{Highly Maintainable}\quad To remain useful, KnowWhereGraph must be easily maintained by the community. This includes both the ease of data integration and how amenable the schema --- and thus the graph --- are to modification: either through the incorporation of new or evolving use cases, rectifying conceptual errors in the graph, or adapting to changes in the data sources.

%%% Solution
% Candidate
As a consequence of these stringent requirements, we identified the Modular Ontology Modeling (MOMo) methodology as a prime candidate for developing the KnowWhereGraph Ontology. 
% MOMo and Pattern-based Methods
MOMo is a \emph{pattern-based} method, meaning that KWG leverages existing ontology design patterns (either developed by the community or extracted from existing standards) \cite{odp1}, modifies them to the appropriate context or use case \cite{template}, and assembles these modules into a modular ``plug-and-play'' schema, allowing for maximum maintainability \cite{modont,momo-swj}. 

\strike{Such a pattern-based approach furthermore allows for relatively straightforward dataset integration.}
Such a pattern-based approach furthermore facilitates dataset integration. When integrating multiple datasets, it can be convenient to conceptualize them along the same (ontological) dimension. For example, understanding tabular data about a place as observations, similarly to how a hazard impacts a place (and the measurement thereof) is also an observation. This results in a predictable method for querying against the final, integrated schema of the knowledge graph.

%%% Key Contributions
Explicitly, the contributions that this manuscript describes are as follows:
\begin{compactitem}
    \item the KnowWhereGraph Ontology (2023), design principles, and implementation;
    \item the principled leveraging of a discrete global grid for spatial integration; and
    \item a representative example of answering (one aspect of) an important question that is driven by one of KWG's use-case scenarios.
\end{compactitem}

%%% Directory
The structure of the paper is as follows. 
First, we provide in Section~\ref{sec:prelim} some preliminary concepts to orient readers.
Section~\ref{sec:appr} details the ontology modeling approach taken to develop the KnowWhereGraph Ontology, which is presented in Section~\ref{sec:desc}, alongside discussion on its implementation. 
In Section~\ref{sec:example}, we provide context on how the ontology can be leveraged to use the KWG.
Section~\ref{sec:related} discusses related work and parallel efforts, and how the KWG connects or overlaps.
Finally, in Section~\ref{sec:conc}, as we conclude, we identify next steps.

%%%%%%%%%%%%%%%%%%%%%%%%%%%%%%%%%%%%%%%%%%%%%%%%%%%%%%%%%%%%%%%%%
\section{Preliminaries}
\label{sec:prelim}
%%%%%
First, we introduce two important concepts: the Discrete Global Grid, which informs some of our ontological choices, and schema diagrams, which use a simple graphical syntax to convey intuitive relationships between classes. Finally, we provide a full directory of availability\add{,} and usage information can be found in Section~\ref{ssec:avail}.

%%%%%%%%%%%%%%%%%%%%%%%%%%%%%%%%%%%%%%%%%%%%%%%%%%%%%%%%%%%%%%%%%
\subsection{Discrete Global Grid}
\label{ssec:dgg}
%%%%
\begin{figure*}
    \centering
    \includegraphics[width=\textwidth]{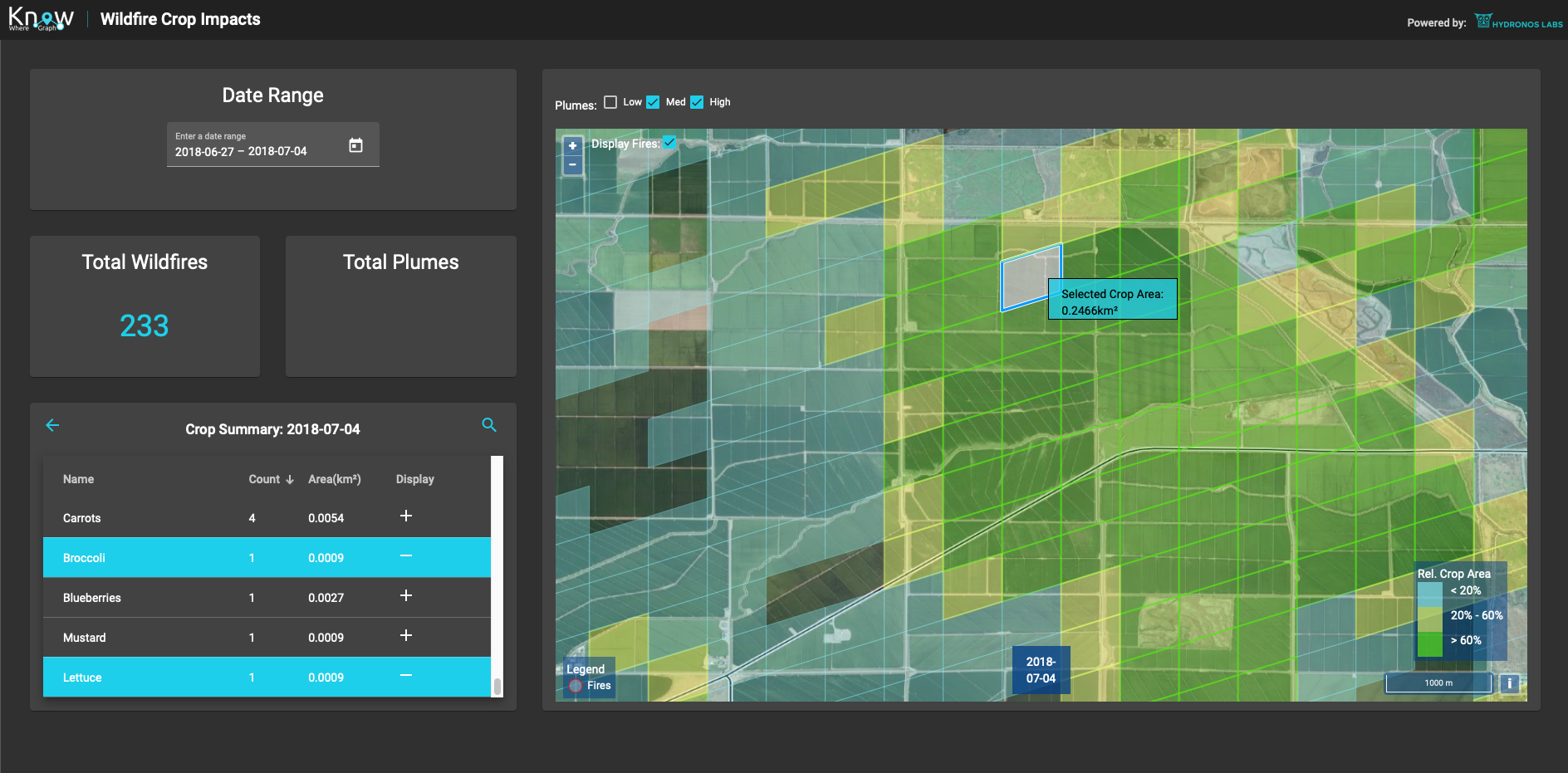}
    \caption{This interface displays a grid-based view of land-use/land-cover data, which are integrated with data from nearby wildfires and their smoke plumes. The grid tessellates the Earth and is called a \emph{discrete global grid}.}
    \label{fig:s2-view}
\end{figure*}
%%%%%
% Opening
One of the primary motivations for KWG is to act as a geospatial informational backbone for any dataset describing physical phenomena and their impacts or places and their characteristics. All such datasets are mapped onto a \emph{discrete global grid} (DGG) \cite{dgg}. In our case, it is a hierarchical DGG, which then acts as the the common spatial data frame to achieve a compromise between data precision (i.e., the fidelity of the geometries underlying geospatial phenomena, including regions), access speed, and ease of data integration among multiple vector and raster \replace{data sets}{datasets}, while supporting globally unique identifiers for the constituent cells.

% Definition (broadly)
In particular, the hierarchical DGG is a partitioning of the surface of the Earth into some number of ``top-level'' parent cells, which are then further partitioned into child cells, and so on until the desired spatial resolution is achieved \cite{dgg,odp-hcf}. 
% Choice
KnowWhereGraph utilizes the S2 Grid System\footnote{\url{https://s2geometry.io/}} implementation, but other DGGs were also considered (e.g., Uber's H3\footnote{\url{https://eng.uber.com/h3/}}).\footnote{A thorough treatment of the different DGG systems is out of scope for this article; we refer the reader to a recent survey \cite{dgg-survey}.}

% Usage
A DGG allows for a common underlying spatial reference system for both geospatial vector and raster datasets, for which we can pre-compute spatial relations between different features or regions for vector data and pre-compute summary statistics for a given cell at a certain level from different raster datasets. By emphasizing the notion of a cell, we can examine exact cell characteristics, predict or infer contents of its parent and child cells, and quickly generate an overview of spatially co-located features and regions of interest without having to compute directly spatial relations.

Furthermore, we can ontologically encode how geospatial information and other phenomena may affect (e.g., transitively) places across the hierarchy of cells within the DGG. This means that we can represent geospatial information and other phenomena at different levels of granularity. This method is discussed more deeply in Section~\ref{sec:desc}, with an example in Section~\ref{sec:example}. Figure~\ref{fig:s2-view} shows an interface that uses the S2 cells to examine quickly regions of the surface of the Earth. By selecting specific cells, we can use them both as a common integrator and as a way to examine what other datasets integrate with (or map onto) specific S2 cells.

%%%%%%%%%%%%%%%%%%%%%%%%%%%%%%%%%%%%%%%%%%%%%%%%%%%%%%%%%%%%%%%%%
\subsection{Schema Diagrams}
\label{ssec:sd}
%%%%%
\replace{Following}{From} the MOMo methodology \cite{momo-swj}, we use a (visual) graphical structure called a schema diagram as our primary method for communicating ontological structure. This diagram carries a reduced semantics; it is meant to be intuitive and easily understood, rather than explicitly and visually conveying the exact, underlying logical axioms. We \replace{do, however, aim for}{use} a consistent visual syntax across all diagrams\replace{.}{, stated below for convenience.} 

Boxes of any non-gray color indicate a class. Goldenrod boxes are atomic classes. \strike{Purple boxes indicate a class that strictly consists of a controlled set of individuals.} Blue boxes (with dashed borders) indicate \emph{hidden complexity} --- i.e., that there are additional relations, but which have been removed from view for clarity. Frequently, this means that it represents a class that is drawn from outside of the KWG namespaces (e.g., OWL Time \cite{time-tr}). Large gray boxes that encapsulate many arrows and boxes indicate a module, meaning that they are conceptually related. Yellow ellipses indicate a datatype. These are generally prefixed with the appropriate namespace, for clarity. \strike{Occasionally, we will use duplicate boxes (i.e., identical labels), which are a convenience, so as to unclutter the diagram of intersecting edges.} Filled arrows indicate a binary relation. If one points to a box, then it is an object property. If one points to an ellipse, it is a data property. Open-face arrows indicate a subclass relation. \strike{Occasionally, use of an open arrow, stylized with an additional mark, indicates that the subject or source of the arrow is an individual of the type that is represented by the box at the target.}

%%%%%%%%%%%%%%%%%%%%%%%%%%%%%%%%%%%%%%%%%%%%%%%%%%%%%%%%%%%%%%%%%
\subsection{Resource Availability}
\label{ssec:avail}
%%%%
We provide multiple types of documentation for the KnowWhereGraph Ontology: visual representations (the schema diagrams), living documentation (generated using \add{WiDoCo} \cite{widoco}), the formalization of the ontology, and a static, technical report (generated using \add{a document generator} \cite{doc-gen}) can all be found in \cite{kwg-widoco}. The KnowWhereGraph, its ontologies, and tools are maintained by the KnowWhereGraph team; more information can be found online.\footnote{\url{https://knowwheregraph.org/}}

%%%%%%%%%%%%%%%%%%%%%%%%%%%%%%%%%%%%%%%%%%%%%%%%%%%%%%%%%%%%%%%%%
\section{The Ontology Modeling Approach}
\label{sec:appr}
%%%%%
% Motivate MOMo
For KWG to be both a success as a project, and a usable resource, the ontology, that would act as its schema, would need to be sustained easily beyond the project's lifetime. Furthermore, it would need to be easily adaptable to new data sources or unexpected changes in data sources to the extent possible both during the initial development and beyond the project's funding period as the pilot (vertical) use cases are transformed. As such, we identified quickly the Modular Ontology Modeling methodology (MOMo; \cite{momo-swj}) as an ideal candidate for performing the development work, since the pattern-based approach both allowed for iterative development, and mitigated long-term maintainability concerns with its resulting modular structure. 

% What actually is MOMo
The MOMo methodology is a combination of top-down and bottom-up approaches.
\emph{Top-down} refers to the fact that MOMo generally assumes that the use cases for the ontology (or knowledge graph) are known at the outset. Then, interactions (competency questions) and important concepts (key notions) are identified and further developed, axiomatized, and serialized. Furthermore, MOMo is a pattern-centric methodology, whereby the use of patterns provides a super-structure for an ontology to follow. \emph{Bottom-up} refers to the emphasis on the data and datasets that are available and can be acquired. Then, by identifying the appearance of important concepts (key notions), additional patterns can be identified.

% Drawbacks
However, because this fast-paced application-oriented effort involved a significant number of people (about 50 students, researchers, software engineers, and domain experts, including non-governmental organization [NGO] and industry representatives), a pure MOMo approach was not entirely feasible. A pure application would have required a more detailed up-front understanding of the major data sources and the application scenarios and needs; these were not realizable within the project timelines, as well as both the rapidly changing data landscape and availability, and the initial knowledge of end-user needs. In summary, we did not have a stable understanding of the requirements (inhibiting the top-down approach) and we did not have a stable understanding of the available data or what exactly those data comprised (inhibiting the bottom-up approach).  

% How exactly was it adapted?
As a result, we agreed on two critical aspects: we would use the DGG as our geospatial backbone and utilize a kernel of the SOSA/SSN ontology (the kernel is discussed in Section~\ref{ssec:kernel} and SOSA/SSN and our implementation in Section~\ref{ssec:impl}). Furthermore, it necessitated that we conduct steps 3-5 of the methodology (as listed in the next section) in parallel, by separate teams, and with repeated iterations to ensure convergence.\footnote{In this case, we take convergence to mean either that our conceptualizations of important terms (i.e., phenomena) were consistent, or that we were using the same implementations for quantities and time, for example.} Indeed, in the beginning stages, we did multiple drafts of both which pieces of the SOSA/SSN ontology we expected to reuse and how they would be used as methods of integration. Key to the eventual convergence were overlapping members between teams, which helped propagate a shared conceptualization, willingness to retrace and redo previous work, and a commitment to our initial minimal requirements.
%%%%%
% \begin{figure}
%     \centering
%     \includegraphics[width=\linewidth]{figures/approach}
%     \caption{This shows a graphical view to the bottom-up and top-down components of the ontology development methodology utilized for KWG.}
%     \label{fig:approach}
% \end{figure}
%%%%%%%%%%%%%%%%%%%%%%%%%%%%%%%%%%%%%%%%%%%%%%%%%%%%%%%%%%%%%%%%%
\subsection{The Steps}
\label{ssec:steps}
%%%%%
\add{In brief, we provide this adapted MOMo approach in nine steps, listed as follows with accompanying commentary.}\strike{Briefly, we provide a short description, with commentary of the adapted MOMo approach, and as shown in 
%Figure~\ref{fig:approach}
provides a complementary, graphical view of the approach}.

% Step 1 %%%%%%%%%%%%%%%%%%%%%%%%%%%%%%%%%%%%%%%%%%%%%%%%%%%%%%%%
\subsubsection{Define the use case.} The first step of MOMo has, in general, three stages, 1) to understand the available data, datasets, and provisioning institutions, as well as \replace{the}{any} originating schemas\strike{(if any)}; 2) to identify \replace{possible}{applicable} vocabularies, ontologies, ontology design pattern repositories, and standards; and 3) to work with domain experts, knowledge engineers, and stakeholders to gain an understanding of \strike{the myriad --- possibly competing ---} perspectives on the use case. \strike{A particular knowledge elicitation methodology is not recommended here, merely ascribing what the outcome of such methodology should be.}
An additional complication results from the fact that KnowWhereGraph has multiple use cases, which fall under its overarching role as a geoinformational integration service (i.e., the area briefings). Understanding how the use cases both interacted and could support each other was a continuous process. We provide an example use case in Section~\ref{ssec:use}.

% Step 2 %%%%%%%%%%%%%%%%%%%%%%%%%%%%%%%%%%%%%%%%%%%%%%%%%%%%%%%%
\subsubsection{Develop competency questions.} \strike{This step focuses on the interaction aspects of a traditional knowledge elicitation step. Essentially,}This step \strike{aims to }narrow\add{s} the \add{use case's} scope \strike{of the use case} by explicitly describing \strike{the} question-answering needs, and thus the expected responses one should receive from a \strike{fully materialized} \replace{knowledge graph}{KG}. A competency question takes the form ``How many wildfires had smoke plumes that impacted leafy greens in California in 2019?'' This demonstrates which datasets \replace{should be related and thus}{are} accessible from queries against the graph, thus guiding both design and validation. Additional examples are provided in Section~\ref{ssec:cqs}.

% Step 3 %%%%%%%%%%%%%%%%%%%%%%%%%%%%%%%%%%%%%%%%%%%%%%%%%%%%%%%%
\subsubsection{Identify key notions.} Generally, \emph{key notions} are the concepts that are central to the knowledge graph. For example, this might mean that either they have high connectivity between other concepts, or they may appear in a high proportion of the competency questions. \replace{However, i}In the case of KWG, the key notions appear at two different levels of abstraction, resulting in the integration of the top-down and bottom-up approaches.
At the outset, we knew \add{some} key notions would exist for the entire KnowWhereGraph --- which we call the KWG Core (i.e., Hazards --- events that happen and have the potential to impact people, Regions --- places that are important to people, and S2Cell --- which abstracts our geospatial backbone). We provide more details in Section~\ref{ssec:kwgc}.
Key notions also exist on the level of datasets, and, furthermore, we wanted to use SOSA/SSN \cite{sosa-tr}, to unify how the datasets that describe Phenomena would be ontologized.\footnote{By ontologized, we mean the act of integrating a dataset by providing an appropriate ontological description.}
Essentially, this meant that these first three steps would need to happen in conjunction, while the rest occurred in sequence (although iteratively). New patterns --- and modules --- needed to be developed to describe the KWG Core concepts, a reusable pattern needed to be extracted from SOSA/SSN, and the datasets needed to be ontologized.
This resulted in a clear process for rapidly integrating newly encountered data, leveraging a powerful and robust W3C recommendation, and reusing a well understood pattern.

% Step 4 %%%%%%%%%%%%%%%%%%%%%%%%%%%%%%%%%%%%%%%%%%%%%%%%%%%%%%%%
\subsubsection{Match ODPs to key notions.} Ontology design patterns \cite{odp1} can be considered to be tiny ontologies that solve domain-invariant modeling problems in (purposefully) abstract ways\strike{--- these are sometimes called \emph{conceptual components}}; their purpose is to be adapted to specific use cases. This step matches the conceptual component of a pattern to the key notion in the use case. 
\replace{These p}{P}atterns can be sourced from resources such as the ODP portal \cite{odp-portal} or MODL \cite{modl}. In some cases these pattern repositories may not be sufficient, and it may be necessary to develop new patterns. For example, we needed to develop novel patterns for representing our DGG \cite{odp-hcf}, causal relations between events \cite{odp-ce}, and the alignment of terms between taxonomies \cite{odp-ta}\replace{. We d}{,as discuss}\add{ed} \strike{these in more detail} in Section~\ref{ssec:patterns}.

% Step 5 %%%%%%%%%%%%%%%%%%%%%%%%%%%%%%%%%%%%%%%%%%%%%%%%%%%%%%%%
\subsubsection{Instantiate the patterns to create modules.} This step is the process by which a general or abstract pattern is taken and modified to fit the use case\replace{. The result of these modifications, we call a module. This process is called}{, creating modules through} \emph{template-based instantiation} \cite{template}\strike{, as it differs from other pattern-adaptation strategies insofar that the pattern is not treated as a top-level ontology and its classes subsequently subclassed. Instead, the structure of the pattern is reused}. Intuitively, we can consider the pattern to be a sort of ``fill in the blanks.'' This strategy is \strike{extremely} effective in the case of KWG, as we have multiple datasets that describe physical phenomena and their relation to space and time. By reusing the same pattern for these concepts, we already achieve a baseline conceptual compatibility and interoperability.

% Step 6 %%%%%%%%%%%%%%%%%%%%%%%%%%%%%%%%%%%%%%%%%%%%%%%%%%%%%%%%
\subsubsection{Systematically axiomatize each module.} This is the step where most ontological analysis occurs. In general, the MOMo approach emphasizes the use of \emph{schema diagrams} to communicate the connectivity of data (i.e., it purposefully obscures the \emph{exact} underlying formalization) until this step, where domain experts and knowledge engineers both dig into the specifics of the relations. However, in many cases, since we have chosen a pattern-centric methodology, we could reuse the axiomatization \replace{from}{directly provided by} the pattern. We \replace{provide}{give} an excerpted example \strike{of the axiomatization of a dataset} in Section~\ref{ssec:ex}.

% Step 7 %%%%%%%%%%%%%%%%%%%%%%%%%%%%%%%%%%%%%%%%%%%%%%%%%%%%%%%%
\subsubsection{Assemble the modules.} This is a quick step that involves connecting the different modules together\replace{. Additionally, we sought to add any}{ and seeks to identify} additional axioms \strike{or relations} that would span multiple modules. For example, in KWG, we have explored including causal relations between the events and phenomena of the different datasets. \add{This also includes mappings between different datasets that describe the same phenomenon (e.g., wildfires sourced from the MTBS vs. the NIFC.}

% Step 8 %%%%%%%%%%%%%%%%%%%%%%%%%%%%%%%%%%%%%%%%%%%%%%%%%%%%%%%%
\subsubsection{Review the final product.} 
\strike{This step generally pertains to ensuring the naming consistency, catching any possible URI minting errors, and checking the consistency (in a logical sense) of the assembled modules.}

% Step 9 %%%%%%%%%%%%%%%%%%%%%%%%%%%%%%%%%%%%%%%%%%%%%%%%%%%%%%%%
\subsubsection{Produce artifacts.} 
\strike{This step is the serialization of the ontology (i.e., the assembled modules), generation of the documentation, and polishing of the schema diagrams. Please see Section~
%\ref{ssec:avail}
for the locations of these artifacts.}

\subsubsection{Validation.}
\add{This is not strictly a final step, but a continuous step over many iterations. The ontology was developed in close-contact with the subject matter experts, adjusting for the use-cases, and identifying appropriate competency questions and prototypical SPARQL queries. Additionally, we developed in conjunction with the ontology a set of SHACL shapes to validate data according to the ontology \cite{kwg-shacl}.}

%%%%%%%%%%%%%%%%%%%%%%%%%%%%%%%%%%%%%%%%%%%%%%%%%%%%%%%%%%%%%%%%%
\section{The KnowWhereGraph Ontology}
\label{sec:desc}
%%%%%
The KnowWhereGraph (KWG) Ontology integrates over thirty datasets across multiple spatial and temporal resolutions and thematic dimensions, which are shown in the Appendix (Tables~\ref{tab:tds} and \ref{tab:pds}). By \emph{thematic}, we mean datasets that describe physical phenomena and their relation to a place and time, as well as the measured or observed characteristics of a place or the populations within them (i.e., a county's adult obesity rate). This may be a natural hazard, forecasts, or man made features (e.g., highways). On the other hand, \emph{place-centric} datasets describe human-meaningful places, such as the boundaries of sociopolitical regions.
% Large ontology, how do we describe in an archival way?
Due to the size of the ontology (there are 312 classes and 2980 axioms), it is less convenient to discuss exhaustively each class and their respective axioms. The purpose of this paper is to provide a brief, but thorough holistic overview of the KWG ontology.

To achieve this, we \textbf{(a)} do not explicitly examine our formalization, instead providing pointers to the formalization of each constituent piece; \textbf{(b)} make use of schema diagrams to convey generalized relationships between classes in a simplified visual format\footnote{Note our syntax and motivation in Section~\ref{ssec:sd}.}; and \textbf{(c)} differentiate between our \emph{conceptualization} and \emph{implementation} of the ontology, as a way to specifically address design decisions independent of any reused ontologies.

\begin{figure}[t]
    \centering
    \includegraphics[width=\linewidth]{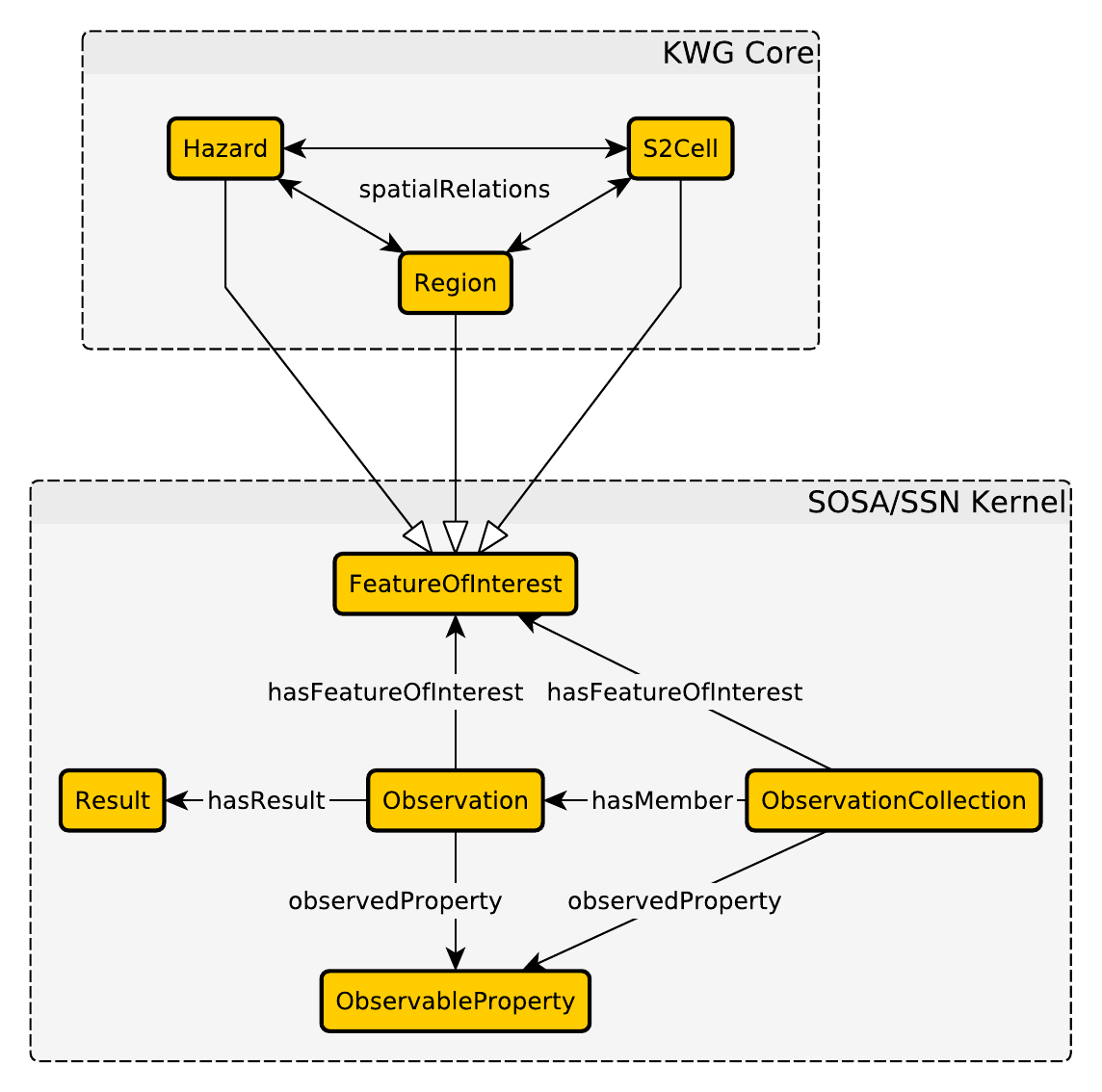}
    \caption{This figure shows the KWG kernel, which consists of three classes and their spatial relations (taken from the KWG Ontology), combined with the SOSA/SSN kernel. \strike{The boxes indicate classes. The grey dashed boxes are for visual partitioning of the kernels. Filled arrows indicate object properties. Open arrows indicate subclass relations.} The \textsf{spatialRelations} label is a placeholder for the KWG Ontology's implementation of the Simple Functions relation ontology.}
    \label{fig:core-kernel}
\end{figure}

%%% Conceptualization
The conceptual layer is discussed in \replace{s}{S}ections~\ref{ssec:kwgc} and \ref{ssec:kernel}: the KWG Core, a set of three tightly related classes, and a reusable kernel whose structure was extracted from (and still eventually implemented using) the SOSA/SSN ontology. 
% Pointer to figure
These are shown in the top and bottom, respectively, in Figure~\ref{fig:core-kernel}. 
%%% Implementation Layer
The implementation of the KWG ontology (i.e., how we materialized these core components) is discussed in Section~\ref{ssec:impl}.
%%% Directory for the rest of the section
In Section \ref{ssec:patterns}, we briefly refer to the ontology design patterns utilized within the KWG Ontology.
Section~\ref{ssec:impl} discusses at a high level the implementations of different components of the KWG Ontology, including metadata, observations, quantities, and time.
Finally, Section~\ref{ssec:kwgon} provides pointers to tangential, yet important, standalone ontologies, which were developed to aid KWG in achieving its objectives.

%%%%%%%%%%%%%%%%%%%%%%%%%%%%%%%%%%%%%%%%%%%%%%%%%%%%%%%%%%%%%%%%%
\subsection{The KnowWhereGraph Core Structure}
\label{ssec:kwgc}
%%%%%
%%%%%
% Opening
The core of the KWG ontology is actually quite simple, as it serves a straightforward purpose: ``hazards need to be linked to the places that they impact.'' These links are served in two ways: direct integration and alignment to the underlying DGG. Thus arise our three main classes: \textsf{Hazard}, \textsf{Region}, and \textsf{Cell}.

% Descriptions of the Classes

% Hazard %%%%%%%%%%%%%%%
\textbf{The \textsf{Hazard} Class}\quad is a specific form of physical phenomena that primarily interests (at the time of this writing) KWG. We consider a hazard to be a physical phenomenon that can (generally negatively) impact civilization in some way and that is anchored in space and time. That is, it will have its own underlying geometry, which is integrated with both the individual cells of the DGG and the underlying geometries (when available) for human-meaningful regions. The \textsf{Hazard} is then temporally scoped in a manner appropriate to the originating dataset. \textsf{Hazard}s and their effects are described via observations; more detail is provided in the next section. The instances of this class (and its subclasses) are generally populated from the thematic datasets, as shown in Table~\ref{tab:tds}.

% Region %%%%%%%%%%%%%%%
\textbf{The \textsf{Region} Class} is the conceptualization of a human-meaningful place for the KWG Ontology. Crucially, they may, or may not, have an explicit geometry. The instances of this class are generally populated from the place-centric Datasets, as shown in Table~\ref{tab:pds}. A \textsf{Region} might have socio-political boundaries (e.g., a US census tract or county) or have natural boundaries (e.g., a river). \textsf{Region}s are aligned to the underlying DGG, providing a backbone to see how arbitrary \textsf{Hazard}s and \textsf{Region}s interact without explicitly performing the spatial integration between the two datasets.

% Cell   %%%%%%%%%%%%%%%
\textbf{The \textsf{Cell} Class} is purposefully generalized and intended to be organized into a DGG. We specifically use the \textsf{S2Cell} class, which is axiomatically specified as  \emph{hierarchical}, via different spatial relations (e.g., RCC-8 \cite{rcc8}). This general class is included in case we wish to support additional DGGs in the future. The purpose of the DGG is to provide a level of interoperability beyond our own knowledge graph. Any phenomenological dataset with a geospatial component can thus be aligned to the \textsf{S2Cell} DGG and immediately achieve interoperability and spatial integration with KWG and its ontology.

As a \textsf{Cell} itself does not rightly exist (e.g., the spatial geometry of the \textsf{Cell} is not human meaningful), but is merely an arbitrary human convenience, any \textsf{Cell} points to a serialized geometry (more detail in Section~\ref{sssec:space}). \textsf{Region}s and \textsf{Hazard}s then overlap, touch, or contain (among the other RCC-8 relations), with the \textsf{S2Cell}s.

%%%%%%%%%%%%%%%%%%%%%%%%%%%%%%%%%%%%%%%%%%%%%%%%%%%%%%%%%%%%%%%%%
\subsection{A Reusable SOSA/SSN Kernel}
\label{ssec:kernel}
%%%%%
% form the underlying semantic backbone for a majority of datasets in KWG and for the KWG ontology as a whole. 
There are two types of integration that KWG supports. The KWG Core classes (from the previous section) integrate across broader notions of space, time, phenomena, and their alignment to some DGG. The second method is through a consistent representation of \emph{how} the different hazards and other geospatial phenomena impact these places (e.g., through storm tracks) or what characteristics can be measured or observed about these places (e.g., demographic statistics). To accomplish this, we extracted a kernel structure from the SOSA/SSN ontology \cite{sosa-tr,sosa-jws,ssn-swj}. Eventually this is implemented directly from the classes of the SOSA/SSN ontology (Sec.~\ref{sssec:obs}), GeoSPARQL (Sec.~\ref{sssec:space}), and QUDT (Sec.~\ref{sssec:qudt}).

We chose this route, in large part, to simplify the process of data integration, as well as eventually leverage W3C and Open Geospatial Consortium (OGC) recommendations and standards, but not utilize every aspect of the more comprehensive ontologies. Furthermore, by using this structure, and in particular the tripartite purposeful disconnect between an observation, the target of the observation (i.e., the feature of interest) and the observed property, allows for quick semantic harmonization between terms. For example, KWG has multiple datasets that describe wildfires and their properties. Yet, not all properties are present across all datasets. However, it is still easy and straightforward to receive all observations about a particular fire without knowing exactly from which dataset the fire originated.

This structure thus forms the underlying semantic backbone of the KWG and for the ontology as a whole; it is shown in the bottom of Figure~\ref{fig:core-kernel} (the green boxes).

%%%%%%%%%%%%%%%%%%%%%%%%%%%%%%%%%%%%%%%%%%%%%%%%%%%%%%%%%%%%%%%%%
\subsection{Utilized Ontology Design Patterns}
\label{ssec:patterns}
%%%%%
During the development of the KWG ontology, it was necessary both to create new patterns and to adapt existing patterns. So far, four new patterns have been developed: the hierarchical features ODP \cite{odp-hcf}, the causal relations ODP \cite{odp-ce}, the taxonomy alignment ODP \cite{odp-ta}, and the computational observation ODP \cite{odp-co}. Existing patterns were selected from MODL \cite{modl}: entity with provenance ODP and the agent role ODP. 

%%%%%%%%%%%%%%%%%%%%%%%%%%%%%%%%%%%%%%%%%%%%%%%%%%%%%%%%%%%%%%%%%
\subsection{Implementation}
\label{ssec:impl}
%%%%%
We have developed several standalone ontologies and resources, and reused quite a few well-known, standardized, or W3C recommended vocabularies, taxonomies, and ontologies, \add{as described below}. This helps to enhance interoperability with other knowledge graphs, to maintain consistency in our data model, and to leverage existing tools that support these vocabularies.

\add{The KnowWhereGraph Ontology is specified in OWL-DL. This allows us to take advantage of scoped and qualified domain and range axioms, as well as axiomatically specified role-chains, transitivity, and functionality using the axiom patterns as found in \cite{owlaxax}. At the time of this writing, there are 150 classes, 70 object properties, and 75 data properties.}

%%%%%%%%%%%%%%%%%%%%%%%%%%%%%%%%%%%%%%%%%%%%%%%%%%%%%%%%%%%%%%%%%
\subsubsection{Modeling and Representing Space}
\label{sssec:space}
%%%%%
We used GeoSPARQL\cite{geosparql-tr,geosparql-swj}, an OGC standard, to represent our geospatial data in RDF. It, in turn, reuses the OGC Simple Features (SF) standard, which defines a set of geometric primitives (e.g., points or polygons) and their spatial relationships. Within the KWG Ontology, we represent any discrete geographic feature type (\textsf{Hazard}, \textsf{Region} and their subclasses) that has a spatial extent as a subclass of GeoSPARQL's \textsf{geo:\strike{Spatial}Feature} class. We also use the spatial relationships from GeoSPARQL (based on DE-9IM spatial relations) to establish pre-computed spatial relationships between any two spatial features. Several graph databases support GeoSPARQL \cite{li2022performance}, including GraphDB \cite{graphdb}, which we use for KWG.

%%%%%%%%%%%%%%%%%%%%%%%%%%%%%%%%%%%%%%%%%%%%%%%%%%%%%%%%%%%%%%%%%
\subsubsection{Modeling Time}
\label{sssec:time}
%%%%%
The Time Ontology is (re)used for all representations of time within the KWG Ontology. The super-property for most time-related conceptualizations is \textsf{kwg-ont:hasTemporalScope}, which effectively has a range of any \textsf{Temporal\-Entity} from OWL:Time and is most often used with \textsf{Hazard}. However, we also specify \textsf{Temporal\-Entity} to be the range of \textsf{resultTime} and \textsf{phenomenon\-Time} for \textsf{Observations}. For serialization of the actual time-related data, we reuse the XML schema datatypes (e.g., \textsf{xsd:dateTime}).

%%%%%%%%%%%%%%%%%%%%%%%%%%%%%%%%%%%%%%%%%%%%%%%%%%%%%%%%%%%%%%%%%
\subsubsection{Modeling Observations}
\label{sssec:obs}
%%%%%
As mentioned previously, SOSA/SSN was reused to implement the \textsf{Observation} and related classes within the KWG ontology. In particular, we model observations made by sensors that detect physical properties (e.g., climate, weather, hazard properties) and chemical properties (e.g., air pollutants), and information about demographics and public health (e.g., disease statistics, socioeconomic indicators such as poverty and social vulnerability). We have accordingly refined the interpretation of two concepts: \textsf{sosa:FeatureOfInterest} and \textsf{sosa:Observation}. 

The first, \textsf{sosa:FeatureOfInterest}, is the main class for representing FOIs (objects and phenomena that are being observed or measured) in the ontology. FOIs can be physical entities or conceptual entities, such as a hazard event, an administrative region boundary, or an abstract concept like S2 cell geometry. The second, \textsf{sosa:Observation}, is a class to represent any measurement of a property of a FOI that is made by a sensor. \textsf{sosa:ObservationCollection} denotes a group of related observations that are made by one or more sensors over a period of time, and share common characteristics such as the observed property or time of observation, and so on. The \textsf{sosa:hasMember} property denotes the set of observations that belong to the collection. An observation or observation collection has a number of properties that describe its characteristics, such as \textsf{sosa:observedProperty}, which indicates the property or phenomenon being observed; \textsf{sosa:madeBySensor} indicates the sensor that made the observation; \textsf{sosa:resultTime} indicates the time at which the observation was made; \textsf{sosa:hasResult} indicates the value of the observation; and \textsf{hasFeatureOfInterest} indicates the entity being observed.

In KWG, a \textsf{sosa:FeatureOfInterest} \replace{represents}{generally satisfies} both (i) \emph{\replace{the}{that it is a} thing whose property can be estimated or calculated} \cite{sosa-tr,sosa-jws} and (ii) \emph{\replace{anything}{that it is a thing} that can have a spatial representation and an associated geometry} \cite{geosparql-tr,geosparql-swj}. In this sense, even though upper ontologies classify events and things/objects/features as distinct classes based on the endurant--perdurant categorization \cite{mascardi2007comparison}, the fact that a hazard satisfies both (i) and (ii) allows us to identify the \textsf{HazardEvent} class both as a \textsf{sosa:FeatureOfInterest} and a \textsf{geo:Feature}. \add{Meanwhile, it is also worth noting that such a refinement (particularly the added definition of (ii)) still aligns with SOSA/SSN's definition of \textsf{sosa:FeatureOfInterest} as in KWG, most datasets are about environmental observations, which are anchored to a spatial representation on the surface of the earth.}

Observations (and their collections) in SOSA are defined as \emph{estimates or calculations of the value of a property made using a sensor (device, agent, or software)}, \strike{while a feature of interest is a \emph{thing whose property is being estimated or calculated in the course of an observation to arrive at a result}}. Although measurements such as census data from community surveys and soil-type maps constructed from surveys conducted over the course of a century do not precisely satisfy this definition, it seems plausible for us to model them using the \textsf{sosa:Observation|ObservationCollection} classes. In this case the features of interest are either some spatial region (\textsf{Region}) such as an administrative region or a region that shares a common property (e.g., soil type), or a spatial grid (\textsf{S2Cell}).

\add{Finally, we would like to highlight that different from traditional works of using SOSA/SSN, in which \textsf{sosa:observableProperty} (often defined by experts) is used to integrate different datasets, KWG mainly relies on the spatiotemporal dimension (e.g., \textsf{Region} or \textsf{Cell}) of the observation to achieve the integration, and our way of modeling \textsf{sosa:observableProperty} is mainly bottom-up starting from the nature of the real world datasets. Employing such a strategy is due to the fact that KWG attempts to integrate over 30 thematic datasets and their observable properties are mostly heterogeneous and there is a lack of standards to harmonize them semantically. Instead, spatial and temporal information are relatively easier to be standardized, and every environmental observation happens at some location during some period of time.}

%%%%%%%%%%%%%%%%%%%%%%%%%%%%%%%%%%%%%%%%%%%%%%%%%%%%%%%%%%%%%%%%%
\subsubsection{Modeling the KWG Metadata}
%%%%%
As KWG and its ontology grew, it became very difficult to understand the contents of the graph because its size engendered project management gaps (i.e., overlapping graph development efforts due to similarity in phenomena or diverging sources of datasets as new, better, or less stale sources were identified). Furthermore, we wished to ascribe provenance to the constituent data, but avoiding so-called triple explosion.\footnote{Triple explosion is what we call the process of compact (tabular) data becoming a disproportionate number of triples (i.e., subject-predicate-object statements). In this case, we wished to avoid every \textsf{Observation} in our graph carrying detailed provenance, when identical data would be repeated for every observation from a dataset.} 
% To address these problems, we developed the metadata model (as shown in Figure~\ref{fig:metadata}) that tracks this information. 
\add{To do so, we mainly mark the provenance of observable properties because they are modeled differently in different datasets. Specifically,} this strategy links observable properties to the datasets from which they were derived --- and the organizations that originally provided them --- as opposed to a per assertional statement ascription of provenance. While this slightly increased query complexity (i.e., the number of statements between the original provenance and any particular observation is increased), this slight increase in complexity actually created a much clearer top-level view of the graph and what it contains. This strategy is distinct from creating named graph, as we are not directly encapsulating any triples into a distinct graph. Instead, we are explicitly grouping the \textsf{ObservableProperty}s and \textsf{FeatureOfInterest}s, which can \strike{be} then be used to extract a subgraph, as necessary.

The metadata model distinguishes between two different types of subgraphs that appear in the KWG: \emph{dataset} and \emph{thematic}. A dataset subgraph is straightforward: it is merely an encapsulation of a particular dataset as materialized against the KWG Ontology. A thematic subgraph is a set of dataset subgraphs that are unified according to some theme --- which for now is merely modeled as a \textsf{xsd:string}. Our thematic examples largely pertain to our pilot use cases: land valuation (which incorporates the dataset subgraphs for smoke plumes, smoke plume forecasts, wildfires, and logistics) and disaster response (which incorporates subgraphs for hurricane impact predictions, FEMA disaster declarations, and public health indicators). Every dataset subgraph is spatially integrated against our different regional representations.

Additionally, this metadata model shored up project management of large graphs by allowing for a representation of the KnowWhereGraph team itself, and it indicates who was in charge of developing particular subgraphs. 

We reused several vocabularies to annotate the KWG Ontology. For instance we used annotation properties from Dublin Core Metadata Initiative (DCMI) Metadata Terms \cite{weibel1998dublin} to describe the title, description, rights, license, date created and creator. We used Friend of a Friend (FOAF\footnote{\url{http://xmlns.com/foaf/0.1/}}) to describe the development team and their roles. The Simple Knowledge Organization System (SKOS) was used to annotate definitions, examples, and the taxonomic structure between domain concepts. Finally, we used PROV-O \cite{provo-tr} and the Data Catalog Vocabulary \cite{dcat-tr} to connect the subgraphs to the datasets from which they were derived.

%%%%%%%%%%%%%%%%%%%%%%%%%%%%%%%%%%%%%%%%%%%%%%%%%%%%%%%%%%%%%%%%%
\subsubsection{Modeling Quantities}
\label{sssec:qudt}
%%%%%
\begin{figure}[t]
    \centering
    \includegraphics[width=\linewidth]{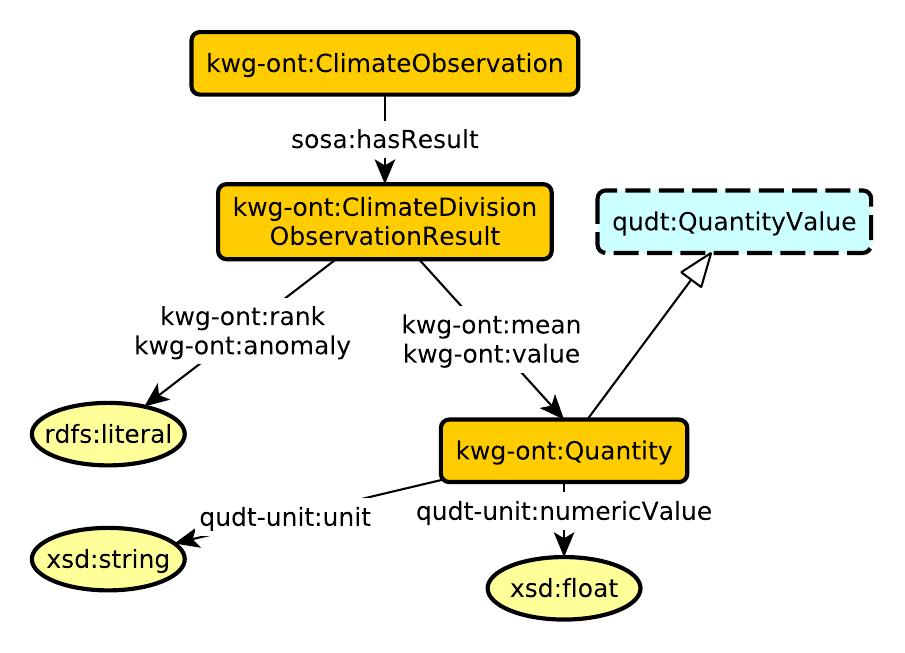}
    \caption{This schema diagram shows how QUDT is used for representing measured quantities, their values and units of measurement within KWG.}
    \label{fig:qudt}
\end{figure}
%%%%% 
The QUDT (Quantities, Units, Dimensions and Data Types) ontology \cite{rijgersberg2013ontology} is used for representing climate measurement data (such as temperature, Palmer drought severity index, cooling degree days) and their corresponding units of measure. Specific climate quantity types (such as mean or value) are denoted using the \textsf{kwg-ont:Quantity} class, a subclass of \textsf{kwg-ont:QuantityValue}. Measured values and corresponding data properties are then captured using data properties \textsf{qudt-unit:unit} and \textsf{qudt-unit:numericValue}.

%%%%%%%%%%%%%%%%%%%%%%%%%%%%%%%%%%%%%%%%%%%%%%%%%%%%%%%%%%%%%%%%%
\subsection{The KnowWhereGraph Ontology Network}
\label{ssec:kwgon}
%%%%%
While developing the KWG Ontology, we identified the need for integrating, harmonizing, and reasoning over specific domain data in the graph, and also for consistently representing these data following standard definitions and best practices. Thus, two standalone ontologies were developed that are imported by the KWG Ontology, yet are not interdependent (i.e., other ontologies may play the role\footnote{These ontologies were developed since we were not able to find sufficient other resources to use without doing so.}). We provide brief pointers for reference and documentation on how they assist the KWG Ontology in meeting the requirements of the overall graph.% This was specific for the variety of hazard observation data, and for the expert-expertise data included to support humanitarian relief purposes. Our search for suitable ontologies that could be reused ended up revealing that for these two specific domains there were no established, standardized and well-tested resource readily available. This pushed us to develop the two ontologies discussed below, which are implemented in KWG.

%%%%%%%%%%%%%%%%%%%%%%%%%%%%%%%%%%%%%%%%%%%%%%%%%%%%%%%%%%%%%%%%%
\subsubsection{The Expertise Ontology}
\label{sssec:eo}
%%%%%
\begin{figure}[t]
    \centering
    \includegraphics[width=\linewidth]{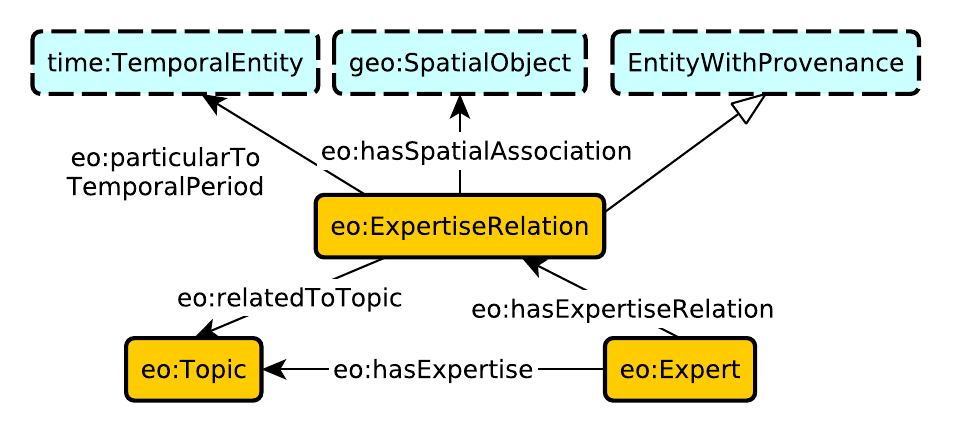}
    \caption{\replace{Simplified}{This} schema diagram \replace{of}{shows a simplified view of} the Expertise Ontology.}
    \label{fig:eo}
\end{figure}
%%%%% 
KWG contains information on agents who are experts on topics related to specific disaster types, disaster management activities, named disasters, and public health. The Expertise Ontology\footnote{\url{https://github.com/KnowWhereGraph/expertise-ontology}} (EO) was developed to represent all varied expertise-related information consistently. At a high level, EO consists of a core set of classes and properties to 1) model experts (people and groups), topics and their relations, 2) represent hierarchical relations between topics of different levels of granularity, and 3) connect topics with relevant content in a KG. EO facilitates representing not only research-based and theory-based expertise, but also experience-based expertise. To do so, EO models the activities that an expert may have engaged in, or their role and affiliation within an organization, and scopes these spatially and temporally.

%%%%%%%%%%%%%%%%%%%%%%%%%%%%%%%%%%%%%%%%%%%%%%%%%%%%%%%%%%%%%%%%%
\subsubsection{The Disaster Management Domain Ontology}
\label{sssec:hdo}
%%%%%
\begin{figure}[t]
    \centering
    \includegraphics[width=\linewidth]{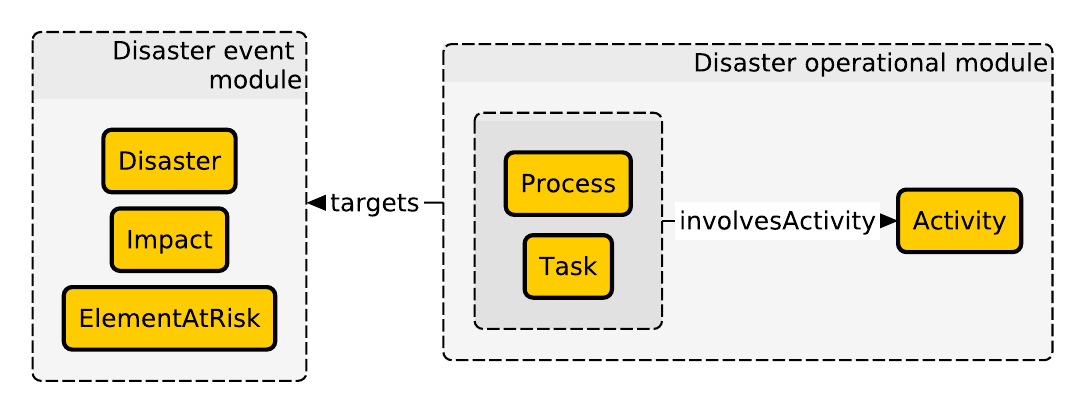}
    \caption{\replace{High-level}{This} schema diagram \strike{of} the DMDO \add{at a high level}. \strike{Orange boxes indicate classes from the \emph{Hazard Event} module. Blue boxes indicate classes from the \emph{Hazard Operational} module.}}
    \label{fig:dmdo}
\end{figure}
%%%%%
KWG contains at least 11 hazard datasets and at least one hazard (\textsf{Fire}) characterized by data from four different sources (see Table~\ref{tab:tds}). To model their semantics and enable integration using existing ontologies, we are developing the Disaster Management Domain Ontology (DMDO), which will provide a framework both to align diverse hazard types, formats of data, and domain vocabularies consistently within KWG, and to clarify better situational awareness of the spatiotemporal interactions of similar events. The ontology disambiguates hazards from disasters and their impacts, but also distinguishes spatiotemporal events from their observations. DMDO also formalizes the United Nations Office for Disaster Risk Reduction (UNDRR) hazard classification \cite{UNDRR_Report} using the taxonomy alignment ODP \cite{odp-ta}. Fig~\ref{fig:dmdo} shows some of the core concepts in each module and their connections. 

%%%%%%%%%%%%%%%%%%%%%%%%%%%%%%%%%%%%%%%%%%%%%%%%%%%%%%%%%%%%%%%%%
\section{Intended Usage \& Example}
\label{sec:example}
%%%%%
In this section, we introduce one of our core use-case scenarios. In brief, a use-case scenario is a narrative coupled with a set of competency questions. Together, these describe an expected set of interactions between an end user and the KWG. Then, we provide a glimpse at what data --- materialized against the KWG Ontology --- look like, and how they can be served to answer those competency questions.

%%%%%%%%%%%%%%%%%%%%%%%%%%%%%%%%%%%%%%%%%%%%%%%%%%%%%%%%%%%%%%%%%
\subsection{Use Case Scenario}
\label{ssec:use}
%%%%
One context in which we envision KWG to be used is humanitarian response. To explore this, we partnered with Direct Relief,\footnote{\url{https://www.directrelief.org/}} a non-profit organization that responds to disasters worldwide by delivering urgently needed medical supplies. The organization's effectiveness depends largely on their ability to quickly and efficiently provide resources that are appropriate for the situation, often in the face of logistical challenges like interrupted communication channels and supply chains. Among their primary tasks following a disaster are to (a) identify local authorities and experts who can advise on immediate medical needs; and (b) assess which supplies are likely in demand --- and where --- based on the nature of the disaster and demographic and health characteristics of the affected populations. As often unpredictable and complex events, disasters challenge humanitarian responders to comprehend and address many issues simultaneously and immediately. From the operational state of a region's hospitals and health clinics to the prevalence of poverty and chronic health disorders within a community, information on a variety of themes can contribute to a responder's situational awareness and ability make informed decisions.

%%%%%%%%%%%%%%%%%%%%%%%%%%%%%%%%%%%%%%%%%%%%%%%%%%%%%%%%%%%%%%%%%
\subsection{Selected Competency Questions}
\label{ssec:cqs}
%%%%%
Typical questions following a hurricane's landfall might include, 
\begin{compactenum}[CQ1.]
	\item\emph{Who is most vulnerable?} 
	\item\emph{What are the critical health risks faced by these populations?} 
	\item\emph{Who has expertise relevant to this situation and place, and how do we contact them?} 
	\item\add{\emph{How can we tailor our response to respect the historical and cultural sensitivities of affected populations?}}
\end{compactenum}

%%%%%%%%%%%%%%%%%%%%%%%%%%%%%%%%%%%%%%%%%%%%%%%%%%%%%%%%%%%%%%%%%
\subsection{Worked Example}
\label{ssec:ex}
%%%%
\begin{figure*}[tbh]
    \centering
    \includegraphics[width=\textwidth]{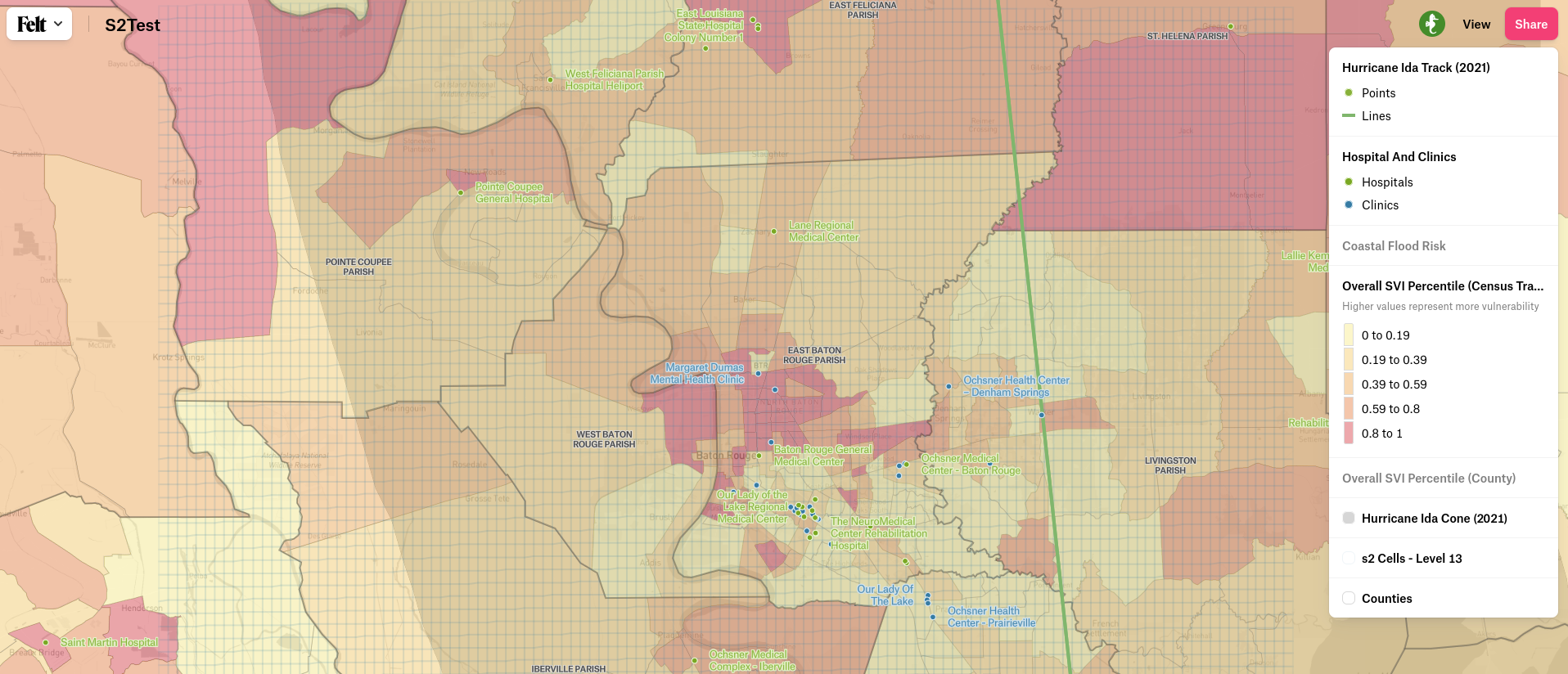}
    % \missingfigure{}
    \caption[]{This figure displays a visualization of the geo-enriched data that KWG provides. The yellow-to-red gradient depicts Social Vulnerability Index scores by census tract. Superimposed on this are S2 DGG cells and boundaries of the counties (parishes) of the Baton Rouge, LA, USA area. Finally, the areal extent of Hurricane Ida's impacts is projected around the track across the landscape (the grey shading and green line, respectively). This provides a quick and intuitive grasp of where attention might be most valuable when no other information is available (e.g., early in the disaster-response scenario).\footnotemark}
    % \url{test}}
    \label{fig:vuln}
\end{figure*}
\footnotetext{This visualization is powered by \url{https://felt.com/}.}

\begin{figure}
\begin{framed}
\footnotesize
\begin{verbatim}
SELECT * WHERE {
  ?cell a kwg-ont:S2Cell .
  ?county a kwg-ont:AdminRegion_3 ;
    geo:sfWithin 
      kwgr:Earth.NA.US.USA.19_1 .
  ?cell kwg-ont:sfWithin ?county .
  ?county sosa:isFeatureOfInterestOf ?obs .
  ?obs a kwg-ont:VulnerabilityObservation .
  ?obs sosa:hasSimpleResult ?result .
}
\end{verbatim}
\end{framed}
\caption[]{This query represents a portion of Figure \ref{fig:vuln}. The exact query can be found online,\footnotemark and we note that the full replication would be best served by combining multiple queries and overlaying them in an external tool. This particular component retrieves all of the Social Vulnerability Index values for all counties in Louisiana, USA, as well as all S2 cells that compose the counties.}
\label{fig:query}
\end{figure}

For a brief example, we will \add{first} focus on CQ1 from the previous section. Figure~\ref{fig:vuln} shows how this question would be answered via a geo-enrichment service built on top of KWG. At the basic level, we see the S2 DGG (rendered at level 13, which means about 1.27 square kilometers per grid cell) combined with Administrative Region boundaries at Level 3 (corresponding to counties, parishes, or boroughs) in the Baton Rouge, Louisiana, USA region. The yellow-to-red gradient indicates the values of the Social Vulnerability Index, provided by the CDC\footnote{\url{https://www.atsdr.cdc.gov/placeandhealth/svi/index.html}} and integrated into KWG. The grey shaded area and green line indicate the areal extent of impact and path of Hurricane Ida, respectively. Altogether, these provide a quick, intuitive overview of the \emph{a priori} state of the disaster. That is, when available data are scarce known shortly after landfall, can educated guesses be made regarding where aid will be needed first? Thus, this gives a holistic answer to CQ1.

Figure~\ref{fig:query} gives a fragment of the abbreviated query that would replicate this view.%\footnotetext{}

\begin{figure*}[tbh]
    \centering
    \includegraphics[width=\linewidth]{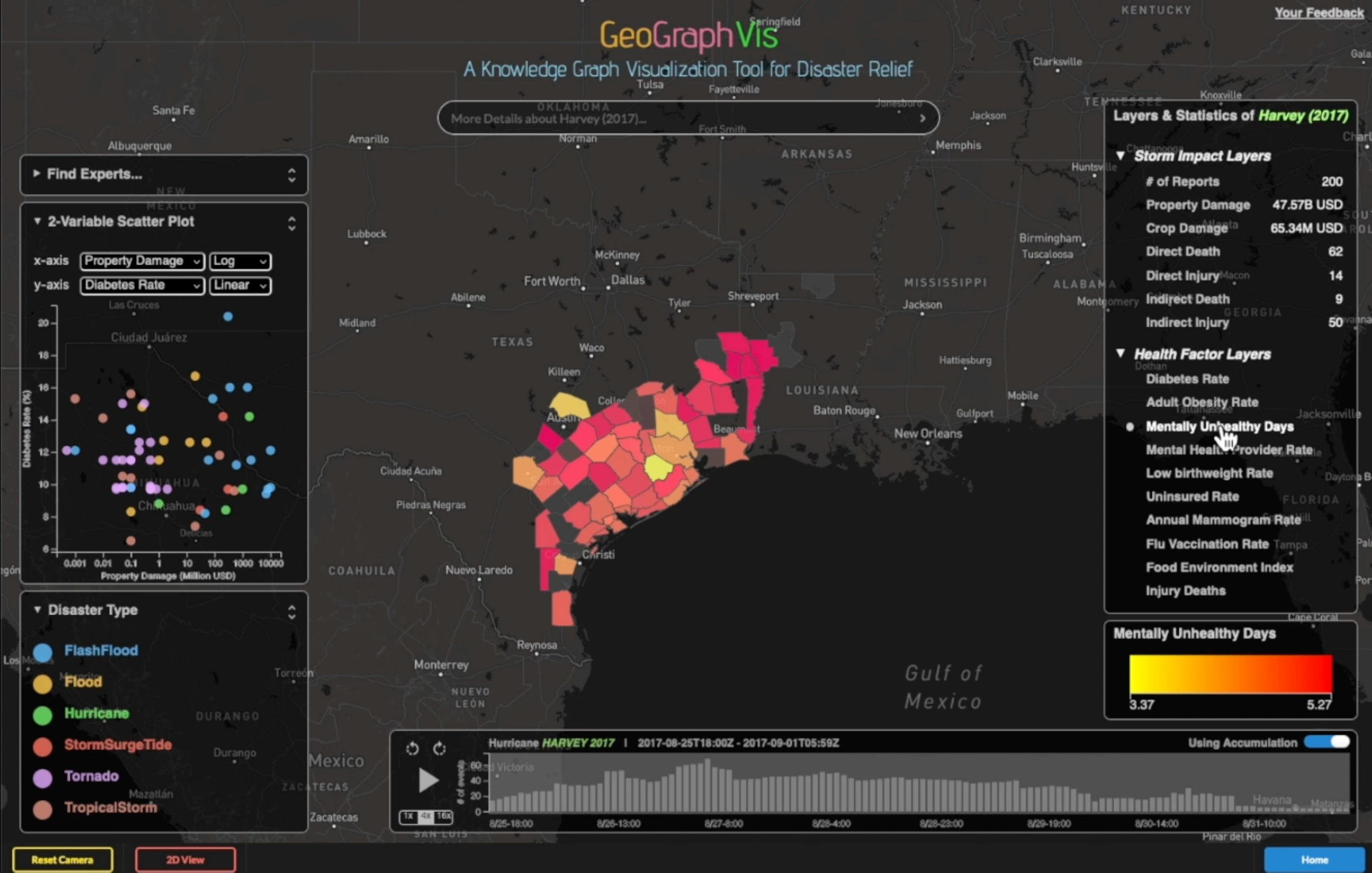}
    \caption{This figure shows the GeoGraphVis \citep{li2023geographvis} visual analytics interface that connects to KnowWhereGraph to support interactive decision-making and enhance disaster management.}
    \label{fig:health_risk}
\end{figure*}

\add{CQ2 is also a critical question for identifying health risks faced by communities vulnerable to disasters. Figure~\ref{fig:health_risk} shows the health factors, including mentally unhealthy days (highlighted in the figure), diabetes rates, adult obesity rates, and injury deaths information from authoritative sources such as the Centers for Disease Control and Prevention and the University of Wisconsin Population Health Institute. These data provide a county-level health profile that helps relief experts identify communities that may encounter more health challenges post-disaster and are in critical need of medical supplies, especially in cases of power outages and infrastructure damage.}

\begin{figure*}[tbh]
    \centering
    \includegraphics[width=\linewidth]{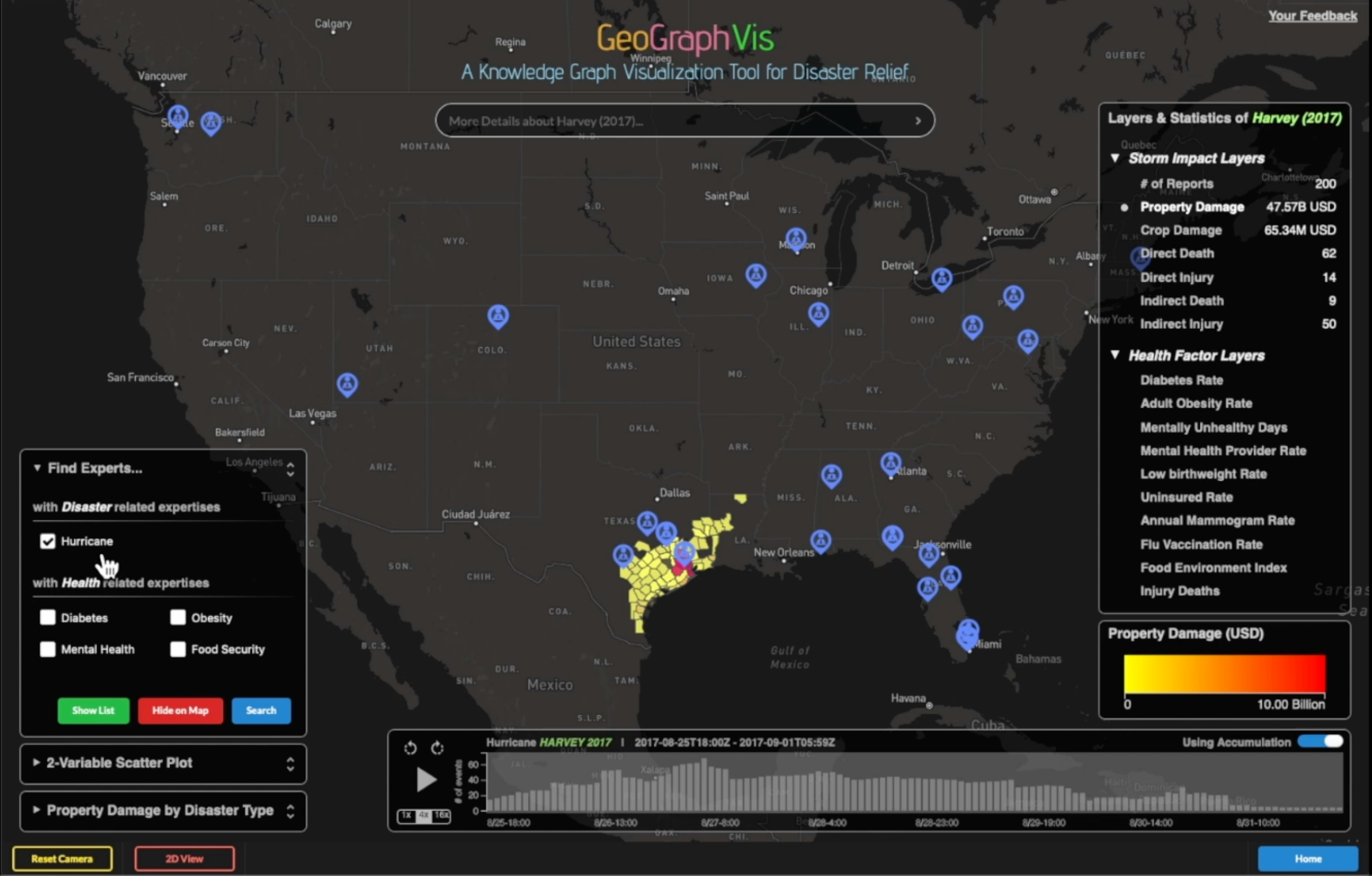}
    \caption{This figure shows the GeoGraphVis \citep{li2023geographvis} visual analytics interface that connects to KnowWhereGraph for searching and identifying local and national experts with desired (disaster response) expertise.}
    \label{fig:whohasexpertise}
\end{figure*}

For any disaster response actions, it is crucial to acquire support from experts who possess both strong domain knowledge (e.g., disaster management) and local knowledge (e.g., medical conditions in local clinics). The answer to CQ3, therefore, offers insights to help decision-makers quickly identify experts with relevant expertise from the KnowWhereGraph. This expert-expertise graph is built by automatically searching and parsing literature (e.g., research publications, reports) using machine learning and large language models. Figure~\ref{fig:whohasexpertise} shows the search results for experts with expertise in ``hurricanes.'' This search criterion can be combined with other factors, such as public health-related expertise, to find both local and national experts who could provide specialized knowledge to guide the distribution of medical supplies in response to disasters.

%%%%%%%%%%%%%%%%%%%%%%%%%%%%%%%%%%%%%%%%%%%%%%%%%%%%%%%%%%%%%%%%%
\section{Related Work}
\label{sec:related}
%%%%%
At the time of this writing, we are not aware of an ontology that performs hazard (or geospatial phenomenological) data and regional identifier integration over a common geospatial backbone (i.e., the S2 DGG) while maintaining a rich description of the phenomena, their impacts, provenance, and lineage. Of course, we are aware of several ontologies that can be leveraged to accomplish this (and we have described many of them in Section~\ref{ssec:impl}). 
\add{However, there seems to be some unproven but very persistent claims in some ontology modeling circles that there be significant benefit in re-using existing ontologies whenever possible. We disagree with this premise, and have found no scientific, published evidence for it, or evidence that would lay out in which cases it is beneficial and in which it is not.}
Briefly, we discuss such ontologies that we encountered but did not re-use, as well as graphs that fill a similar niche to KWG. 

% The SOSA/SSN has been used as standard de-facto for modelling sensors and observations for disciplines such as environmental \cite{roussey2020weather}, agricultural \cite{bakker2021common}, traffic \cite{klotz2018vsso}, and IoT \cite{elsaleh2020iot}. Moreover SOSA is designed to be compatible with various upper ontologies, such as Dolce-Ultralite, but additionally also with other W3C standards such as O\&M, OBOE, and PROV-O \cite{haller2019modular}
\subsection{Environmental Concepts}
\textbf{The Environment Ontology} (ENVO) \cite{buttigieg2013environment} was determined to have limitations that impacted its usability and effectiveness within the context of KWG's use cases. For instance, while ENVO covers a wide range of environmental concepts, it has limited coverage of human-related environmental concepts, such as environmental pollution. Some definitions in ENVO \replace{are ambiguous or imprecise, which led to consistent misinterpretation}{were difficult to reuse}. For example, the term `flood' as a `continuant' versus `flooding' as an `occurent' made it somewhat ambiguous for us to categorize a flood, as reported by NOAA. Further, their usage of an upper ontology and its principles made it difficult to adequately perform the necessary integration over significantly heterogeneous datasets and sources. Due to this strong ontological commitment, it was determined that trying to extend ENVO in a new direction was outside of the scope effort for both KWG and ENVO.

\textbf{SWEET} (Semantic Web for Earth and Environmental Terminology) \cite{raskin2005knowledge} has several overlapping definitions and inconsistent use of relations. For example, `phenomenon' and `observable property' refer to measurable or observable characteristics of the natural world in their definitions. \strike{Yet another example is where `part of' and `has part' are used interchangeably to describe hierarchical relationships between concepts.} Moreover, these vocabularies are large, complex\add{, and have an unclear strategy for their maintenance}, with many concepts and relationships, which makes it challenging to understand and use them effectively. \add{Our focus, however, was on a set of lightweight ontologies and patterns that model all layers from an observational perspective, focusing on measurement properties and less on class membership of the phenomena.}

\subsection{Similar Resources \add{\& Frameworks}}
\textbf{The Google Data Commons}\footnote{\url{https://www.datacommons.org/}} integrates many public datasets and, indeed, draws inspiration from the Google Knowledge Panels and (humbly) our own interfaces. The data integration occurs through Schema.org annotations. Thus, the composite schema is quite shallow. Additionally the graph itself is not made public, but can be accessed through a variety of rate-limited APIs.

\textbf{Esri's ArcGIS Knowledge} is an enterprise knowledge graph specifically built to be compatible with their GeoEnrichment platform, ArcGIS Knowledge Server. It integrates and makes accessible through the platform a set of curated data, as well as common publicly available or governmental datasets. We are not aware of an associated ontology with their graph structure. Additionally, this resource was released during the development of KWG and its ontology.

\add{
\textbf{RDF Data Cube Vocabulary} is a mechanism for modeling multi-dimensional data, while linking data sets and concepts. It was considered early in the development stages of KWG, however, due to difficulties in appropriately aligning the spatial and phenomenological datasets in an intuitive manner, we opted for the SOSA/SSN model, instead. 
}

\add{
\textbf{LinkedGeoData} is a project that aims to add a spatial dimension to the Linked Data and Semantic Web communities. It does this by providing a programmatic framework for converting OpenStreetMaps\footnote{\url{https://openstreetmaps.org/}} to RDF. LinkedGeoData is an excellent tool, but is limited in its ability to represent phenomological data. However, it is targeted as a valuable addition to a future release of KWG, especially to complement our already integrated USDOT\footnote{\url{https://transportation.gov/}} dataset.
}
%%%%%%%%%%%%%%%%%%%%%%%%%%%%%%%%%%%%%%%%%%%%%%%%%%%%%%%%%%%%%%%%%
\section{Conclusion}
\label{sec:conc}
%%%%%
The KnowWhereGraph is a complex project with multiple evolving use cases, a large team, and an ambitious goal. To create the schema for the underlying knowledge graph --- the KnowWhereGraph Ontology --- it was necessary to choose a methodology that produced a schema where emphasizing sustainability and adaptability were prioritized. To this effect, we chose --- and adapted --- the Modular Ontology Modeling methodology, to produce a pattern-based (modular) ontology that could quickly incorporate new data sources, and be easily maintained beyond the project lifetime. Additionally, it was executed in a distributed fashion, resulting in a quickly changing schema, but has since converged. 

As a result, we have developed the KnowWhereGraph Ontology and reported on its kernel structure (i.e., the top-level, repeated structure) of our knowledge graph, the ontology design patterns we have utilized, and a description of the metadata model used to manage and record the provenance of our graph. 

%%%%%%%%%%%%%%%%%%%%%%%%%%%%%%%%%%%%%%%%%%%%%%%%%%%%%%%%%%%%%%%%%
\section*{Future Work}
%%%%%
Over the course of developing the KnowWhereGraph Ontology, we have identified many opportunities for next steps, as follows.
\begin{compactitem}
    \item Develop tighter integration with our developed patterns, in particular for the support of identifying causality between events.
    \item Extract additional patterns that would be useful in the development of other spatially enabled knowledge graphs, such as a simple pattern for representing entities with geometries or a pattern for representing observations.
    \item \add{Create additional mappings to external vocabularies, such as those mentioned in Section~\ref{sec:related}.}
\end{compactitem}

%%%%%%%%%%%%%%%%%%%%%%%%%%%%%%%%%%%%%%%%%%%%%%%%%%%%%%%%%%%%%%%%%
\noindent\emph{Acknowledgement.}   
This work was funded by the National Science Foundation under Grant 2033521 A1: KnowWhereGraph: Enriching and Linking Cross-Domain Knowledge Graphs using Spatially-Explicit AI Technologies. Any opinions, findings, and conclusions or recommendations expressed in this material are those of the authors and do not necessarily reflect the views of the National Science Foundation.

%%%%%%%%%%%%%%%%%%%%%%%%%%%%%%%%%%%%%%%%%%%%%%%%%%%%%%%%%%%%%%%%%
%%%%%%%%%%%%%%%%%%%%%%%%%%%%%%%%%%%%%%%%%%%%%%%%%%%%%%%%%%%%%%%%%
\bibliographystyle{abbrv}
\bibliography{refs}

\begin{thebibliography}{10}

\bibitem{geosparql-swj}
R.~Battle and D.~Kolas.
\newblock Enabling the geospatial semantic web with parliament and geosparql.
\newblock {\em Semantic Web}, 3(4):355--370, 2012.

\bibitem{dgg-survey}
B.~Bondaruk, S.~A. Roberts, and C.~Robertson.
\newblock Assessing the state of the art in discrete global grid systems: Ogc
  criteria and present functionality.
\newblock {\em Geomatica}, 74(1):9--30, 2020.

\bibitem{buttigieg2013environment}
P.~L. Buttigieg, N.~Morrison, B.~Smith, C.~J. Mungall, S.~E. Lewis, and
  E.~Consortium.
\newblock The environment ontology: contextualising biological and biomedical
  entities.
\newblock {\em Journal of biomedical semantics}, 4:1--9, 2013.

\bibitem{rcc8}
A.~G. Cohn, B.~Bennett, J.~Gooday, and N.~M. Gotts.
\newblock Qualitative spatial representation and reasoning with the region
  connection calculus.
\newblock {\em GeoInformatica}, 1(3):275--316, 1997.

\bibitem{dcat-tr}
S.~Cox, R.~Albertoni, A.~G. Beltran, D.~Browning, A.~Perego, and P.~Winstanley.
\newblock Data catalog vocabulary ({DCAT}) - version 3.
\newblock {W3C} recommendation, W3C, Aug. 2024.
\newblock https://www.w3.org/TR/2024/REC-vocab-dcat-3-20240822/.

\bibitem{time-tr}
S.~Cox and C.~Little, editors.
\newblock {\em {Time Ontology in OWL}}.
\newblock {W3C} Recommendation 19 October 2017, 2017.
\newblock Available from https://www.w3.org/TR/2017/REC-owl-time-20171019/.

\bibitem{owlaxax}
A.~Eberhart, C.~Shimizu, S.~Chowdhury, M.~K. Sarker, and P.~Hitzler.
\newblock Most of {OWL} is rarely needed.
\newblock In {\em 18th ESWC}, 2020.
\newblock Under review.

\bibitem{odp1}
A.~Gangemi and V.~Presutti.
\newblock Ontology design patterns.
\newblock In S.~Staab and R.~Studer, editors, {\em Handbook on Ontologies},
  International Handbooks on Information Systems, pages 221--243. Springer,
  2009.

\bibitem{widoco}
D.~Garijo.
\newblock {WIDOCO:} {A} wizard for documenting ontologies.
\newblock In C.~d'Amato, M.~Fern{\'{a}}ndez, V.~A.~M. Tamma,
  F.~L{\'{e}}cu{\'{e}}, P.~Cudr{\'{e}}{-}Mauroux, J.~F. Sequeda, C.~Lange, and
  J.~Heflin, editors, {\em The Semantic Web - {ISWC} 2017 - 16th International
  Semantic Web Conference, Vienna, Austria, October 21-25, 2017, Proceedings,
  Part {II}}, volume 10588 of {\em Lecture Notes in Computer Science}, pages
  94--102. Springer, 2017.

\bibitem{dgg}
M.~F. Goodchild.
\newblock Discrete global grids for digital earth.
\newblock In {\em International Conference on Discrete Global Grids}, pages
  26--28. Citeseer, 2000.

\bibitem{graphdb}
Graph{DB}.
\newblock \url{http://graphdb.ontotext.com/}.

\bibitem{ssn-swj}
A.~Haller, K.~Janowicz, S.~J.~D. Cox, M.~Lefran{\c{c}}ois, K.~Taylor, D.~L.
  Phuoc, J.~Lieberman, R.~Garc{\'{\i}}a{-}Castro, R.~Atkinson, and C.~Stadler.
\newblock The modular {SSN} ontology: {A} joint {W3C} and {OGC} standard
  specifying the semantics of sensors, observations, sampling, and actuation.
\newblock {\em Semantic Web}, 10(1):9--32, 2019.

\bibitem{template}
K.~Hammar and V.~Presutti.
\newblock Template-based content {ODP} instantiation.
\newblock In K.~Hammar, P.~Hitzler, A.~Krisnadhi, A.~Lawrynowicz, A.~G.
  Nuzzolese, and M.~Solanki, editors, {\em Advances in Ontology Design and
  Patterns [revised and extended versions of the papers presented at the 7th
  edition of the Workshop on Ontology and Semantic Web Patterns, WOP@ISWC 2016,
  Kobe, Japan, 18th October 2016]}, volume~32 of {\em Studies on the Semantic
  Web}, pages 1--13. {IOS} Press, 2016.

\bibitem{modont}
P.~Hitzler and C.~Shimizu.
\newblock Modular ontologies as a bridge between human conceptualization and
  data.
\newblock In P.~Chapman, D.~Endres, and N.~Pernelle, editors, {\em Graph-Based
  Representation and Reasoning - 23rd International Conference on Conceptual
  Structures, {ICCS} 2018, Edinburgh, UK, June 20-22, 2018, Proceedings},
  volume 10872 of {\em Lecture Notes in Computer Science}, pages 3--6.
  Springer, 2018.

\bibitem{sosa-tr}
K.~Janowicz, A.~Haller, S.~Cox, M.~Lefran\c{c}ois, D.~L. Phuoc, and K.~Taylor.
\newblock Semantic sensor network ontology.
\newblock {W3C} recommendation, W3C, Oct. 2017.
\newblock https://www.w3.org/TR/2017/REC-vocab-ssn-20171019/.

\bibitem{sosa-jws}
K.~Janowicz, A.~Haller, S.~J.~D. Cox, D.~L. Phuoc, and M.~Lefran{\c{c}}ois.
\newblock {SOSA:} {A} lightweight ontology for sensors, observations, samples,
  and actuators.
\newblock {\em J. Web Semant.}, 56:1--10, 2019.

\bibitem{kwg-aimag-22}
K.~Janowicz, P.~Hitzler, W.~Li, D.~Rehberger, M.~Schildhauer, R.~Zhu,
  C.~Shimizu, C.~K. Fisher, L.~Cai, G.~Mai, J.~Zalewski, L.~Zhou, S.~Stephen,
  S.~G. Estrecha, B.~D. Mecum, A.~Lopez{-}Carr, A.~Schroeder, D.~Smith, D.~J.
  Wright, S.~Wang, Y.~Tian, Z.~Liu, M.~Shi, A.~D'Onofrio, Z.~Gu, and
  K.~Currier.
\newblock Know, know where, knowwheregraph: {A} densely connected, cross-domain
  knowledge graph and geo-enrichment service stack for applications in
  environmental intelligence.
\newblock {\em {AI} Mag.}, 43(1):30--39, 2022.

\bibitem{diverse-data}
K.~Janowicz, C.~Shimizu, P.~Hitzler, G.~Mai, S.~Stephen, R.~Zhu, L.~Cai,
  L.~Zhou, M.~Schildhauer, Z.~Liu, Z.~Wang, and M.~Shi.
\newblock Diverse data! diverse schemata?
\newblock {\em Semantic Web}, 13(1):1--3, 2022.

\bibitem{geosparql-tr}
F.~Knibbe, J.~Herring, J.~Abhayaratna, M.~Bonduel, M.~Perry, N.~J. Car,
  S.~J.~D. Cox, and T.~Homburg.
\newblock Geosparql ontology.
\newblock {W3C} recommendation, W3C, Oct. 2021.
\newblock
  https://opengeospatial.github.io/ogc-geosparql/geosparql11/index.html.

\bibitem{li2023geographvis}
W.~Li, S.~Wang, X.~Chen, Y.~Tian, Z.~Gu, A.~Lopez-Carr, A.~Schroeder,
  K.~Currier, M.~Schildhauer, and R.~Zhu.
\newblock Geographvis: a knowledge graph and geovisualization empowered
  cyberinfrastructure to support disaster response and humanitarian aid.
\newblock {\em ISPRS International Journal of Geo-Information}, 12(3):112,
  2023.

\bibitem{li2022performance}
W.~Li, S.~Wang, S.~Wu, Z.~Gu, and Y.~Tian.
\newblock Performance benchmark on semantic web repositories for spatially
  explicit knowledge graph applications.
\newblock {\em Computers, environment and urban systems}, 98:101884, 2022.

\bibitem{mascardi2007comparison}
V.~Mascardi, V.~Cord{\`\i}, and P.~Rosso.
\newblock A comparison of upper ontologies.
\newblock In {\em Woa}, volume 2007, pages 55--64, 2007.

\bibitem{odp-portal}
Ontology design patterns . org (odp).
\newblock https://ontologydesignpatterns.org/.

\bibitem{raskin2005knowledge}
R.~G. Raskin and M.~J. Pan.
\newblock Knowledge representation in the semantic web for earth and
  environmental terminology (sweet).
\newblock {\em Computers \& geosciences}, 31(9):1119--1125, 2005.

\bibitem{rijgersberg2013ontology}
H.~Rijgersberg, M.~Van~Assem, and J.~Top.
\newblock Ontology of units of measure and related concepts.
\newblock {\em Semantic Web}, 4(1):3--13, 2013.

\bibitem{provo-tr}
S.~Sahoo, D.~McGuinness, and T.~Lebo.
\newblock {PROV}-o: The {PROV} ontology.
\newblock {W3C} recommendation, W3C, Apr. 2013.
\newblock http://www.w3.org/TR/2013/REC-prov-o-20130430/.

\bibitem{momo-swj}
C.~Shimizu, K.~Hammar, and P.~Hitzler.
\newblock Modular ontology modeling.
\newblock {\em Semantic Web}, 14(3):459--489, 2023.

\bibitem{modl}
C.~Shimizu, Q.~Hirt, and P.~Hitzler.
\newblock {MODL:} {A} modular ontology design library.
\newblock In K.~Janowicz, A.~A. Krisnadhi, M.~P. Villal{\'{o}}n, K.~Hammar, and
  C.~Shimizu, editors, {\em Proceedings of the 10th Workshop on Ontology Design
  and Patterns {(WOP} 2019) co-located with 18th International Semantic Web
  Conference {(ISWC} 2019), Auckland, New Zealand, October 27, 2019}, volume
  2459 of {\em {CEUR} Workshop Proceedings}, pages 47--58. CEUR-WS.org, 2019.

\bibitem{doc-gen}
C.~Shimizu and P.~Hitzler.
\newblock Automatically generating human readable documentation for ontology
  design patterns.
\newblock In O.~Seneviratne, C.~Pesquita, J.~Sequeda, and L.~Etcheverry,
  editors, {\em Proceedings of the {ISWC} 2021 Posters, Demos and Industry
  Tracks: From Novel Ideas to Industrial Practice co-located with 20th
  International Semantic Web Conference {(ISWC} 2021), Virtual Conference,
  October 24-28, 2021}, volume 2980 of {\em {CEUR} Workshop Proceedings}.
  CEUR-WS.org, 2021.

\bibitem{odp-co}
C.~Shimizu, P.~Hitzler, and C.~F.~V. II.
\newblock A pattern for modeling computational observations.
\newblock In V.~Sv{\'{a}}tek, V.~A. Carriero, M.~Poveda{-}Villal{\'{o}}n,
  C.~Kindermann, and L.~Zhou, editors, {\em Proceedings of the 13th Workshop on
  Ontology Design and Patterns {(WOP} 2022) co-located with the 21th
  International Semantic Web Conference {(ISWC} 2022), Online, October 24,
  2022}, volume 3352 of {\em {CEUR} Workshop Proceedings}. CEUR-WS.org, 2022.

\bibitem{odp-hcf}
C.~Shimizu, R.~Zhu, G.~Mai, C.~K. Fisher, L.~Cai, M.~Schildhauer, K.~Janowicz,
  P.~Hitzler, L.~Zhou, and S.~Stephen.
\newblock A pattern for features on a hierarchical spatial grid.
\newblock In {\em IJCKG'21: The 10th International Joint Conference on
  Knowledge Graphs, Virtual Event, Thailand, December 6 - 8, 2021}, pages
  108--114. {ACM}, 2021.

\bibitem{odp-ce}
C.~Shimizu, R.~Zhu, M.~Schildhauer, K.~Janowicz, and P.~Hitzler.
\newblock A pattern for modeling causal relations between events.
\newblock In {\em Proceedings of the 12th Workshop on Ontology Design and
  Patterns (WOP 2021), co-located with the 20th International Semantic Web
  Conference (ISWC 2021) : online, October 24, 2021}, volume 3011, pages
  38--50, 2021.

\bibitem{odp-ta}
S.~Stephen, C.~Shimizu, M.~Schildhauer, R.~Zhu, K.~Janowicz, and P.~Hitzler.
\newblock A pattern for representing scientific taxonomies.
\newblock In V.~Sv{\'{a}}tek, V.~A. Carriero, M.~Poveda{-}Villal{\'{o}}n,
  C.~Kindermann, and L.~Zhou, editors, {\em Proceedings of the 13th Workshop on
  Ontology Design and Patterns {(WOP} 2022) co-located with the 21th
  International Semantic Web Conference {(ISWC} 2022), Online, October 24,
  2022}, volume 3352 of {\em {CEUR} Workshop Proceedings}. CEUR-WS.org, 2022.

\bibitem{kwg-widoco}
The {KnowWhereGraph} schema.
\newblock https://stko-kwg.geog.ucsb.edu/lod/ontology.

\bibitem{UNDRR_Report}
Hazard definition and classification review.
\newblock
  \url{https://www.undrr.org/publication/hazard-definition-and-classification-review},
  2021.

\bibitem{weibel1998dublin}
S.~Weibel, J.~Kunze, C.~Lagoze, and M.~Wolf.
\newblock Dublin core metadata for resource discovery.
\newblock Technical report, OCLC, 1998.

\bibitem{kwg-shacl}
R.~Zhu, C.~Shimizu, S.~Stephen, L.~Zhou, L.~Cai, G.~Mai, K.~Janowicz,
  M.~Schildhauer, and P.~Hitzler.
\newblock {SOSA-SHACL:} shapes constraint for the sensor, observation, sample,
  and actuator ontology.
\newblock In {\em IJCKG'21: The 10th International Joint Conference on
  Knowledge Graphs, Virtual Event, Thailand, December 6 - 8, 2021}, pages
  99--107. {ACM}, 2021.

\end{thebibliography}
%%%%%%%%%%%%%%%%%%%%%%%%%%%%%%%%%%%%%%%%%%%%%%%%%%%%%%%%%%%%%%%%%
\appendix
\section{Tables}
\begin{table*}[tbp]
    \begin{tabularx}{\textwidth}{Y|Y|Y}                                            
Thematic Dataset      & Source Agency          & Example Attributes \\ \hline
Soil Properties       & USDA                   & soil type, farmland class  \\
Wildfires             & USGS, USDA, USFS, NIFC & wildfire type, num acres burned \\
Earthquakes           & USGS                   & magnitude \\
Climate Hazards       & NOAA                   & casulaties, property damage \\
Experts (Covid-19 Mobility) & DR               & name, affiliation, expertise  \\
Expert (General)      & KWG, UCS, DR, SS       & name, affiliation, expertise \\
Cropland Types        & USDA                   & crop types (raster data) \\
Air Quality           & EPA                    & air quality index \\
Smoke Plumes Forecasts & NOAA                  & daily smoke plume forecast \\
Climate               & NOAA                   & temperature, precipitation \\
Disaster Declarations & FEMA                   & area, program, amount approved \\
Smoke Plume Extents   & NOAA                   & smoke plume extent \\
BlueSky Forecasts     & BlueSky                & PM10, PM5 \\
Highway Networks      & DOT                    & road type, road length, signage \\
Public Health Observations   & CDC, USCB, UW:PHI & poverty, diabetes, obesity \\
Public Health Infrastructure & HIFLD           & pharmacies, hospitals \\
Social Vulnerability  & CDC/ATSDR              & social vulnerability index \\
Hurricane Tracks      & NOAA                   & max wind speed, min pressure       
\end{tabularx}%\medskip\\
    \caption{Thematic Datasets}
    \label{tab:tds}
\end{table*}
\begin{table*}[tbp]
    \begin{tabularx}{\textwidth}{Y|Y|Y}
        Place-Centric Dataset  & Defining Authority & Spatial Coverage \\ \hline
        S2 cells               & Google             & Lvl 9 (Global), Lvl 13 (US) \\
        Global Administrative Regions & GADM.org    & Global \\
        US Federal Judicial Districts & DoJ, Esri   & US \\
        National Weather Zones        & NOAA        & US \\
        FIPS Codes                    & USCB        & US \\
        Designated Market Areas       & Nielen      & US \\
        ZIP Codes                     & USPS        & US \\
        Climate Divisions             & NOAA        & US \\
        Census Metropolitan Area      & USCB        & US \\
        Drought Zone                  & NDMC        & US \\
        GNIS                          & USGS        & US 
    \end{tabularx}
    \caption{This table shows the Place-centric Datasets which are integrated in the KWG.}
    \label{tab:pds}
\end{table*}

\begin{table*}[tbp]
    \begin{tabularx}{\textwidth}{Y|Y}
        ATSDR  & Agency for Toxic Substances and Disease Registry \\
        CDC    & United States Center for Disease Control \\
        DHS    & United States Department of Homeland Security \\
        DoJ    & United States Department of Justice \\
        DoT    & United States Department of Transportation \\
        DR     & Direct Relief \\
        EPA    & Environmental Protection Agency \\
        FEMA   & Federal Emergency Management Agency \\
        HIFLD  & Homeland Infrastructure Foundation-Level Data \\
        KWG    & KnowWhereGraph \\
        NDMC   & National Drought Mitigation Center \\
        NIFC   & National Interagency Fire Center \\
        NOAA   & National Oceanographic and Atomospheric Administration \\
        SS     & Semantic Scholar \\
        UCS    & University of California System \\
        UCSB   & University of California, Santa Barbara \\
        USCB   & University States Census Bureau \\
        USDA   & United States Department of Agriculture \\
        USGS   & United States Geological Survey \\
        USPS   & United States Postal Service \\
        UW:PHI & University of Wisconsin: Public Health Institute \\
    \end{tabularx}
    \caption{This table displays the abbreviations that we use within the paper, especially the tables.}
    \label{tab:abbrvs}
\end{table*}

\begin{table*}[tbp]
    \begin{tabularx}{\textwidth}{Y|Y}
        kwg-ont: & \url{http://stko-kwg.geog.ucsb.edu/lod/ontology/} \\
        kwgr: & \url{http://stko-kwg.geog.ucsb.edu/lod/resource/} \\
        sosa: & \url{http://www.w3.org/ns/sosa/}
    \end{tabularx}
    \caption{This table displays the prefixes that we use within Section~\ref{ssec:use}, and more broadly within our graph.}
    \label{tab:pfs}
\end{table*}

%%%%%%%%%%%%%%%%%%%%%%%%%%%%%%%%%%%%%%%%%%%%%%%%%%%%%%%%%%%%%%%%%
\end{document}